\documentclass[DIV=9]{scrartcl}
\usepackage[utf8]{inputenc}

\usepackage[dvipsnames]{xcolor}
\definecolor{color1}{RGB}{230,57,70}
\definecolor{color2}{RGB}{29,53,87}
\definecolor{color3}{RGB}{69,123,157}
\usepackage[colorlinks=true,linkcolor=color1,urlcolor=color2,citecolor=color3,anchorcolor=green,breaklinks=true]{hyperref}

\usepackage{amsthm,amssymb,amsmath,amsfonts,mathtools,dsfont}
\usepackage{newtxtext,newtxmath}
\usepackage{booktabs,multirow}

\input{braket}

\usepackage[capitalize]{cleveref}
\crefname{lemma}{lemma}{lemmas}
\crefname{proposition}{proposition}{propositions}
\crefname{definition}{definition}{definitions}
\crefname{theorem}{theorem}{theorems}
\crefname{conjecture}{conjecture}{conjectures}
\crefname{corollary}{corollary}{corollaries}
\crefname{example}{example}{examples}
\crefname{section}{section}{sections}
\crefname{appendix}{appendix}{appendices}
\crefname{figure}{fig.}{figs.}
\crefname{equation}{eq.}{eqs.}
\crefname{table}{table}{tables}
\crefname{item}{property}{properties}
\crefname{remark}{remark}{remarks}
\crefname{problem}{}{}

\usepackage{subcaption,tikz}
\usetikzlibrary{backgrounds,scopes,quantikz,external}
\tikzset{
    qc/.style={
        row sep={0.75cm,between origins},
        column sep=.40cm
    },
    qctight/.style={
        column sep=.25cm
    },
    ps/.style={
        draw=purple
    }
}
\makeatletter
\newcommand{\cctrl}[1]{
	\vcw{#1}
	\edef\cell{\the\pgfmatrixcurrentrow-\the\pgfmatrixcurrentcolumn}
	\expandafter\pgfutil@g@addto@macro\expandafter\tikzcd@atendlabels\expandafter{%
		\expandafter\latephase@end\expandafter{\cell}
	}
}
\makeatother

\newcommand\field\mathds
\newcommand\op\mathbf

\newcommand\ii{\mathrm i}
\DeclareMathOperator{\BigO}{O}

\usepackage[backend=biber,style=alphabetic,sorting=none,doi=false,isbn=false,url=false,maxbibnames=20,maxcitenames=2]{biblatex}
\bibliography{bibliography.bib,bibliography-2.bib}

\newbibmacro{string+doi}[1]{\iffieldundef{doi}{#1}{\href{https://dx.doi.org/\thefield{doi}}{#1}}}
\DeclareFieldFormat{title}{\usebibmacro{string+doi}{\mkbibemph{#1}}}
\DeclareFieldFormat[article]{title}{\usebibmacro{string+doi}{\mkbibquote{#1}}}
\DeclareFieldFormat[incollection]{title}{\usebibmacro{string+doi}{\mkbibquote{#1}}}

\usepackage{eso-pic,setspace,transparent}
\setkomafont{title}{\transparent{0}}
\setkomafont{author}{\transparent{0}}
\setkomafont{date}{\transparent{0}}

\newcommand\BackgroundPic{
    \put(0,0){
    \parbox[b][\paperheight]{\paperwidth}{
    \includegraphics[width=.9\paperwidth,height=.9\paperheight]{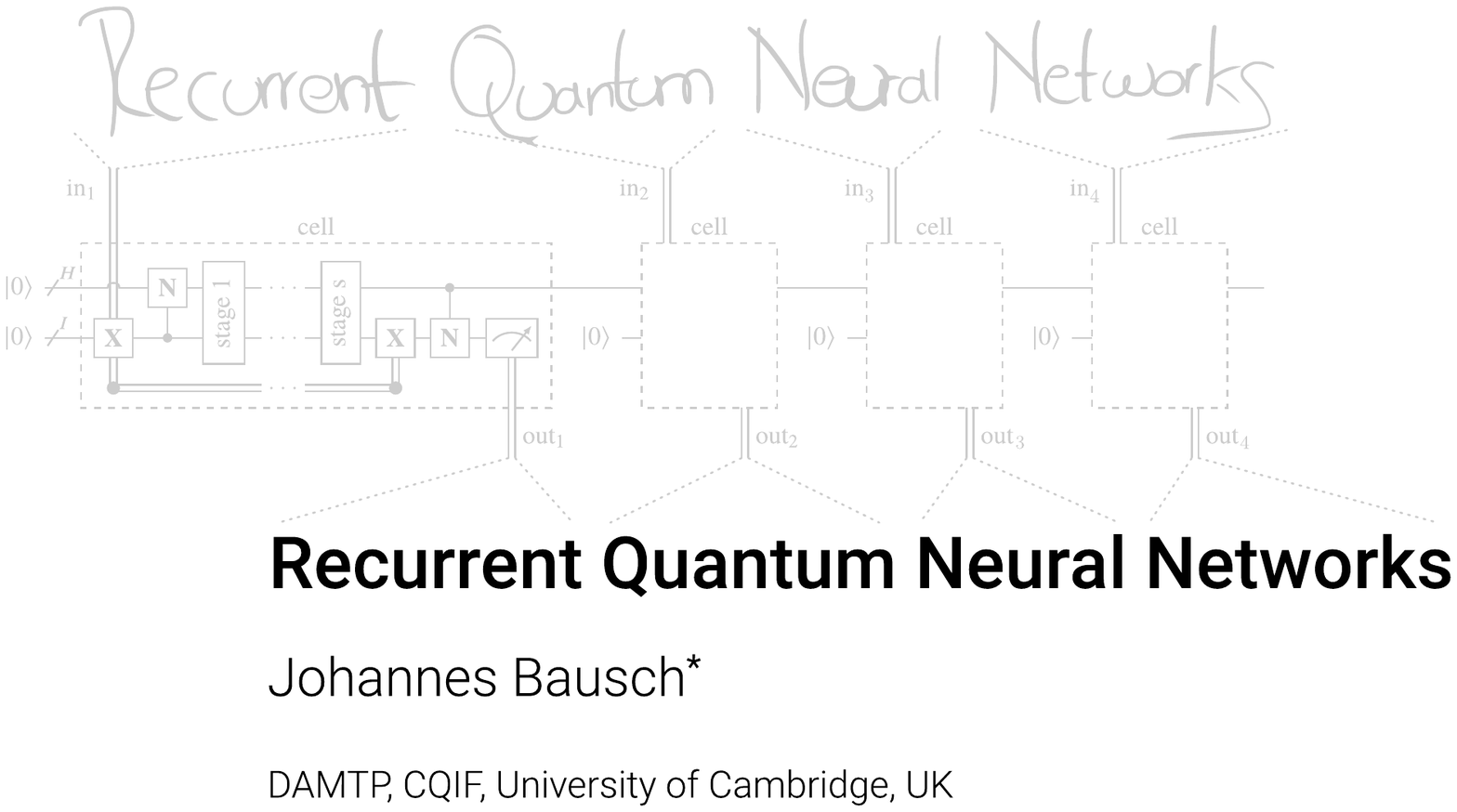}
    \vfill
    }}}
\makeatletter
\newcommand*{\shifttext}[2]{%
  \settowidth{\@tempdima}{#2}%
  \makebox[\@tempdima]{\hspace*{#1}#2}%
}
\makeatother

\title{Recurrent Quantum Neural Networks}
\author{Johannes Bausch\thanks{\texttt{jkrb2@cam.ac.uk}}}
\date{June 2020}

\begin{document}
\AddToShipoutPicture*{\BackgroundPic}
    
\maketitle

\enlargethispage{2cm}
\thispagestyle{empty}

\vspace{3cm}
\begin{abstract}
    \hspace{-1.3cm}\begin{minipage}{12cm}
    \begin{spacing}{1.0}
    Recurrent neural networks are the foundation of many sequence-to-sequence models in machine learning, such as machine translation and speech synthesis. In contrast, applied quantum computing is in its infancy.
    Nevertheless there already exist quantum machine learning models such as variational quantum eigensolvers which have been used successfully e.g.\ in the context of energy minimization tasks.\\[-2mm]
    
    In this work we construct a quantum recurrent neural network (QRNN) with demonstrable performance on non-trivial tasks such as sequence learning and integer digit classification. The QRNN cell is built from parametrized quantum neurons, which, in conjunction with amplitude amplification, create a nonlinear activation of polynomials of its inputs and cell state, and allow the extraction of a probability distribution over predicted classes at each step.\\[-2mm]
    
    To study the model's performance, we provide an implementation in pytorch, which allows the relatively efficient optimization of parametrized quantum circuits with thousands of parameters. We establish a QRNN training setup by benchmarking optimization hyperparameters, and analyse suitable network topologies for simple memorisation and sequence prediction tasks from Elman's seminal paper (1990) on temporal structure learning. We then proceed to evaluate the QRNN on MNIST classification, both by feeding the QRNN each image pixel-by-pixel; and by utilising modern data augmentation as preprocessing step.
    Finally, we analyse to what extent the unitary nature of the network counteracts the vanishing gradient problem that plagues many existing quantum classifiers and classical RNNs.
    \end{spacing}
    \end{minipage}
\end{abstract}

\section{Introduction}
Optimizing recurrent neural networks for long sequences is a challenging task: applying the same RNN cell operator iteratively often suffers from the well-studied vanishing or exploding gradients problem, which results in poor training performance \cite{10.5555/3042817.3043083}.
While long short-term memories or gated recurrent units (LSTMs and GRUs) with their linear operations acting on the cell state have been proposed as a way of circumventing this problem, they too are typically limited to capturing about 200 tokens of context, and e.g.\ start to ignore word order with increasing sequence lengths (more than $50$ tokens away) \cite{Zhu2015a,Dbrowska2008longDistanceDependencies,khandelwal2018LongDistanceLSTM}.

This failure of existing recurrent models to capture very long sequences surely played a role in the advent of alternative, non-recurrent models applicable to sequence-to-sequence tasks, such as transformer-type architectures with self-attention. Yet while these alternatives often outperform LSTMs, they feature a fixed-width context window that does not easily scale with the sequence length; extensions thereof are an active field of research \cite{Al-Rfou2019,Dai2019,kitaev2020reformer}.

Beyond training modifications such as truncated backpropagation \cite{Aicher2019} which attempt to mitigate the vanishing gradient problem for recurrent neural networks directly, there have been numerous proposals to parametrize recurrent models in a way which limits or eliminates gradient decay, by ensuring the transfer operation at each step preserves the gradient norm; examples include orthogonal \cite{Vorontsov2017OnOA} or unitary recurrent neural networks \cite{Arjovsky2015,Widsom16,8668730}.
Yet how to pick an parametrization for the RNN cell that is easy to compute, allows efficient training, and performs well on real-world tasks?
This is a challenging question \cite{Hyland2016}.

In this work, we propose a recurrent neural network model with a parametrization motivated by the emergent field of quantum computation.
The interactions of any quantum system can be described by a Hermitian operator $\op H$, which generates the system's time evolution under the unitary map $\op U(t) = \exp(\ii t \op H)$ as a solution to the Schrödinger equation.
The axioms of quantum mechanics thus dictate that any quantum circuit comprising a sequence of individual unitary quantum gates of the form of $\op U_i(t_i)$---for a set of parameters $t_i$---is intrinsically unitary.
This means that a parametrized quantum circuit serves a prime candidate for a unitary recurrent network.

Such parametrized quantum circuits have already found their way into other realms of quantum machine learning.
A prominent example are variational quantum eigensolvers (VQE), which can serve as a variational ansatz for a quantum state, much akin to how a feed-forward network with a final softmax layer can serve as parametrization for a probability distribution \cite{Wang2019,McClean_2016,Peruzzo2014,Jiang2018}.
VQEs have been deployed successfully e.g.\ in the context of minimizing energy eigenvalue problems within condensed matter physics \cite{Cade2019}, or as generative adversarial networks to load classical probability distributions into a quantum computer \cite{Zoufal2019a}.

To date, classical recurrent models that utilize quantum circuits as sub-routines (i.e.\ where no quantum information is propagated) \cite{Gyongyosi2019}, analysing Hopfield networks on quantum states \cite{Allauddin}, or running classical Hopfield networks by a quantum-accelerated matrix inversion method \cite{Rebentrost2018} have been proposed; yet neither of them have the features we seek: a concrete quantum recurrent neural network with a unitary cell that allows to side-step the problem of gradient decay, and can ideally be implemented and trained on current classical hardware---and potentially on emerging quantum devices in the short-to-mid term.

In this work we construct such a quantum recurrent neural network (QRNN), which features demonstrable performance on real-world tasks such as sequence learning and handwriting recognition.
Its recurrent cell---the operation executed at each step of the input---utilizes a highly-structured parametrized quantum circuits that deviates significantly from those used in the VQE setting.
Its fundamental building block is an improved type of quantum neuron based on \cite{Cao2017} to introduce a nonlinearity, and in conjunction with a type of fixed-point amplitude amplification, allows the introduction of measurements (which are projectors, and not unitary operations) such that the overall evolution on correctly-initialized runs nonetheless remains arbitrarily close to unitary.

With an implementation in pytorch, we repeat several of the learning tasks first proposed in Elman's seminal paper ``Finding Structure in Time'' \cite{Elman1990}, which we utilize to assess suitable model topologies such as the size of the cell state and structure of the parametrized RNN cell, and for benchmarking training hyperparameters for well-established optimizers such as Adam, RMSProp and SGD, as well as more costly methods such as L-BFGS, employed extensively within the context of VQEs.

As a next step, we evaluate the QRNN on integer digit classification, using the standard MNIST dataset.
We find that feeding images pixel-by-pixel allows a discrimination of pairs of digits with up to $99.6\%$ accuracy on the test set.
Using modern data augmentation techniques the QRNN further achieves a test set performance on \emph{all} digits of $\approx 99.2\%$.
In order to demonstrate that the model indeed captures the temporal structure present within the MNIST images, we use the QRNN as a generative model, successfully recreating handwritten digits from an input of \texttt{`0'} or \texttt{`1'}.
As a final experiment, we assess whether the gradient quality decays for long input sequences. In a task of recognizing unique base pairs in a DNA string, we found that even for sequences of 1000 bases---where the target base pairs to be identified are over 500 steps in the past---training performance remains unaffected.

Without question the performance of our proposed model is yet to compete with state-of-the-art scores on popular, much larger datasets than e.g.\ MNIST.
On the other hand, QRNNs are the first quantum machine learning model capable of working with non-superposed training data as high-dimensional as images of integer digits or DNA sequences---meaning binary vectors $v_i \in [0,1]^n$ of length $n>3000$, utilized \emph{not} in quantum superposition as a state $\propto \sum_{i=1}^n v_i \ket{i}$.
By feeding in the vector in a ``one-hot'' fashion, i.e.\ step-by-step as a state $\ket v \ket{v_1}\otimes \ldots \otimes \ket{v_n}$, all information remains accessible to the procedure. Naturally, this would not be within reach to simulate, let alone run on current or near term quantum hardware; feeding $\ket v$ up front into e.g.\ a VQE would require $n$ qubits.

Furthermore, as a variational quantum algorithm capable of being trained with thousands of parameters, benefits such as flattening of local minima emerge; allowing the employment of cheaper, well-established optimization algorithms such as Adam; and enhancing generalization performance \cite{Du2018OnTP,Arora2018OnTO}.

\section{Recurrent Quantum Neural Networks}
\subsection{Parametrized Quantum Gates}
Typical VQE quantum circuits are very dense, in the sense of alternating parametrized single-qubit gates with entangling gates such as controlled-not operations.
This has the advantage of compressing a lot of parameters into a relatively compact circuit.
On the other hand, while it is known that such circuits form a universal family, their high density of entangling gates and lack of connection between parameters results in models that are currently hard to train on classification tasks for inputs larger than a few bits \cite{Benedetti2019}.

In this work, we construct a QRNN cell that is a highly-structured parametrized quantum circuit, built in a fashion such that few parameters are re-utilized over and over, and such that each of them steers a much higher-level logical unit than the components of a VQE circuit.
The cell is built mainly from a novel type of quantum neuron---an extension of \cite{Cao2017}---which rotates its target lane according to a non-linear activation function applied to polynomials of its binary inputs (\cref{eq:eta',eq:neuron-angles}; \cref{fig:activation,fig:qn-1,fig:neuron-rotation} in \cref{sec:qn}).
These neurons are combined in \cref{sec:cell} to form a structured RNN cell, as shown in \cref{fig:qrnn-cell}. The cell is a combination of an input stage that, at each step, writes the current input into the cell state.
This is followed by multiple work stages that compute with access to input and cell state, and a final output stage that creates a probability density over possible predictions.
Applying these QRNN cells iteratively on the input sequence as shown in \cref{fig:qrnn} results in a recurrent model much like traditional RNNs.

During training we perform quantum amplitude amplification (see \cite{Guerreschi2019}) on the output lanes, to ensure that the we measure the correct token from the training data at each step.
While measurements are generally non-unitary operations, the amplitude amplification step ensures that the measurements during training are as close to unitary as we wish.

While the resulting circuits are comparatively deep as compared to a traditional VQE models, they require only as many qubits as the input and cell states are wide (plus a few ancillas for the implementation of quantum neurons and amplitude amplification).

\subsection{A Higher-Degree Quantum Neuron}\label{sec:qn}
The strength of classical neural networks emerges through the application of nonlinear activation functions to the affine transformations on the layers of the network.
In contrast, due to the nature of quantum mechanics, any quantum circuit composed of unitary gates and measurements will necessarily be a linear operation.

\begin{figure}[t]
    \begin{subfigure}[b]{.5\linewidth}
        \newcommand{\qngroup}{\gategroup[3,steps=4,style={dashed,inner xsep=2pt}]{$\op N$}}
        \begin{quantikz}[qc]
        \lstick{$\ket x$} & \qwbundle{n}\qw  & \ctrl1 \qngroup & \qw  & \ctrl1 & \qw & \qw \\
        \lstick{$\ket 0$} & \qw & \gate{\op R(\theta)} & \ctrl1  & \gate{\op R(-\theta)} & \meter[ps]{0} \\
        \lstick{$\ket 0$} & \qw & \qw & \gate{\ii \op Y}  & \qw  & \qw & \qw
        \end{quantikz}
    \end{subfigure}
    \hfill
    \begin{subfigure}[b]{.45\linewidth}
        \begin{quantikz}[qc]
        \lstick{$\ket x$} & \qwbundle{n}\qw  & \gate[3]{\op N} & \qw & \gate[3]{\op N^\dagger} & \qw & \qw  \\
        \lstick{$\ket 0$} & \qw & & \lstick{$\ket 0$} & & & \\
        \lstick{$\ket 0$} & \qw & & \ctrl1 & & \meter[ps]{0}  \\
        \lstick{$\ket 0$} & \qw & \qw & \gate{\ii \op Y} & \qw & \qw  & \qw
        \end{quantikz}
    \end{subfigure}
    \caption{Quantum neuron by \cite{Cao2017}; left a first order neuron, right a second order application. Recursive iteration yields higher-order activation functions, respectively.
    The purple meter indicates a postselection (using fixed-point amplitude amplification) as described in \cite{Tacchino2019}.}
    \label{fig:qn-1}
\end{figure}
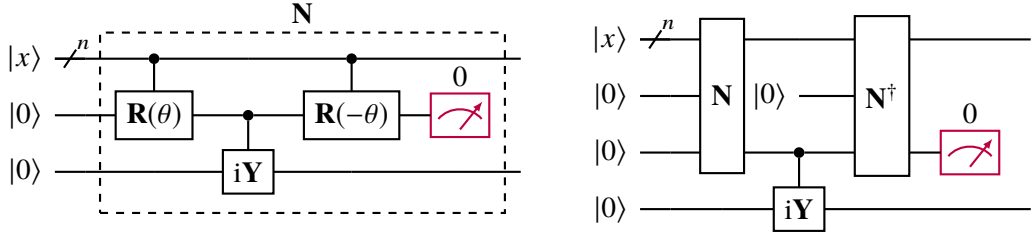

\begin{figure}[t]
    \centering
    \begin{tikzpicture}
    \node[rotate=90] at (-0.5, .25) {$\cos(f(\eta))$};
    \node at (7.7, -2.2) {$=\eta$};
    \node[right] at (0, 0) {\includegraphics[width=9cm]{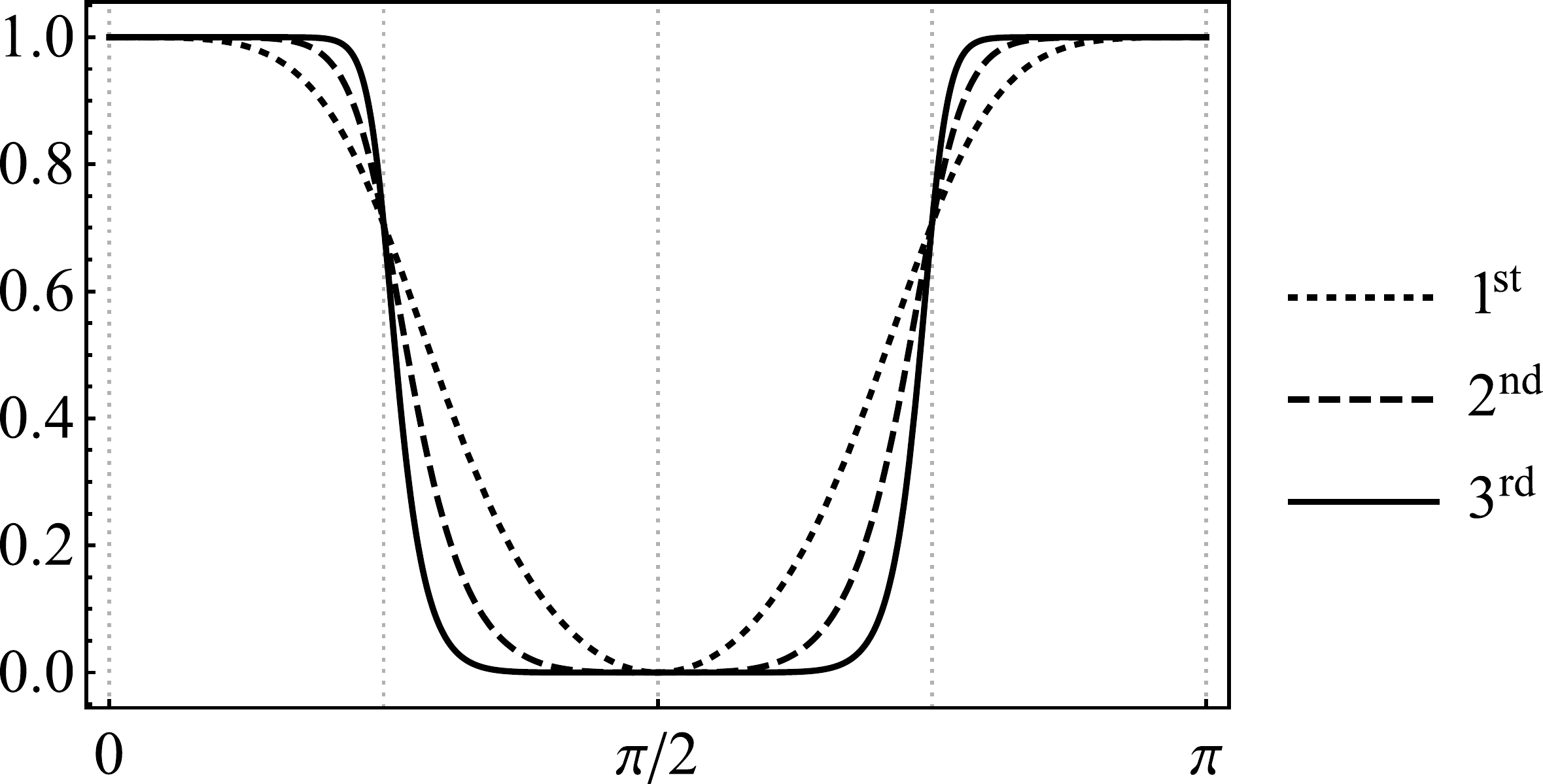}};
    \draw[white] (0,-1.9) -- (1,-1.9);
    \end{tikzpicture}
    \caption{Quantum neuron amplitude $\cos(f(\eta))$, as given in \cref{eq:neuron-angles}.
    Shown are the first to third order activations $o=1,2,3$.}
    \label{fig:activation}
\end{figure}
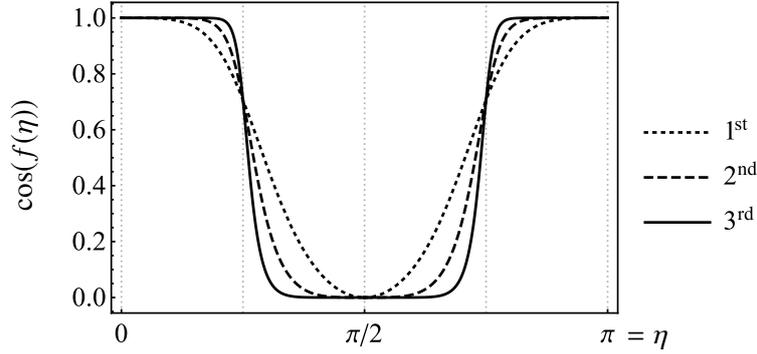

However, this does not mean that no nonlinear behaviour occurs anywhere within quantum mechanics: a simple example is a single-qubit gate $\op R(\theta) := \exp(\ii \op Y \theta)$ for the Pauli matrix $\op Y$, which acts like
\[
\op R(\theta) = \exp\left(\ii \theta \begin{pmatrix} 0 & -\ii \\ \ii & 0 \end{pmatrix} \right) = \begin{pmatrix} \cos\theta & \sin\theta \\ -\sin\theta & \cos\theta \end{pmatrix},
\]
i.e.\ as a rotation within the two-dimensional space spanned by the computational basis vectors of a single qubit, $\{ \ket0, \ket1\}$.
While the rotation matrix itself is clearly a linear operator, we note that the amplitudes of the state---$\cos\theta$ and $\sin\theta$---depend non-linearly on the angle $\theta$.
If we raise the rotation to a controlled operation $\op{cR}(i,\theta_i)$ conditioned on the $i$\textsuperscript{th} qubit of a state $\ket x$ for $x\in\{0, 1\}^n$, one can derive the map
\begin{align}
    \op R(\theta_0) \op{cR}(1,\theta_1)\cdots \op{cR}(n,\theta_n)\ket x \ket0 &= \ket x \big( \cos(\eta) \ket 0 + \sin(\eta) \ket 1 \big) \nonumber\\
    \quad\text{where\ }
    \eta &= \theta_0 + \sum_{i=1}^n \theta_i x_i.\label{eq:eta}
\end{align}
This corresponds to a rotation by an affine transformation of the basis vector $\ket x$ with $x = (x_1, \ldots, x_n) \in \{0, 1\}^n$, by a parameter vector $\theta = (\theta_0,\theta_1, \ldots, \theta_n)$.
This operation extends linearly to superpositions of basis and target states; and due to the form of $\op R(\theta)$ all newly-introduced amplitude changes are real-valued.\footnote{Complex amplitudes are not necessary for the power of quantum computing.
}

\newcommand\ord{\mathrm{ord}}
This cosine transformation of the amplitudes by a controlled operation is already non-linear; yet a sin function is not particularly steep, and also lacks a sufficiently ``flat'' region within which the activation remains constant, as present e.g.\ in a rectified linear unit.
\citeauthor{Cao2017} \cite{Cao2017} proposed a method for implementing a linear map on a set of qubits which yields amplitudes that feature such steeper slopes and plateaus, much like a sigmoidal activation function.
The activation features an order parameter $\ord \ge 1$---the ``order'' of the neuron---that controlls the steepness of the functional dependence, as shown in \cref{fig:activation}; the circuit which gives rise to such an activation amplitude is shown in \cref{fig:qn-1}.

On pure states this quantum neuron gives rise to a rotation by an angle
\[
f(\theta) = \arctan\left( \tan(\theta)^{2^\ord} \right).
\]
Starting from an affine transformation $\eta$ for the input bitstring $x_i$ as given in  \cref{eq:eta}, this rotation translates to the amplitudes
\begin{equation}\label{eq:neuron-angles}
    \cos(f(\eta)) = \frac{1}{\sqrt{1+\tan(\eta)^{2\times 2^\ord}}}
    \quad\text{and}\quad
    \sin(f(\eta)) = \frac{\tan(\eta)^{2^\ord}}{\sqrt{1+\tan(\eta)^{2\times 2^\ord}}},
\end{equation}
emerging from normalising the transformation $\ket 0 \longmapsto \cos(\theta)^{2^\ord}\ket 0 + \sin(\theta)^{2^\ord}\ket 1$, as can be easily verified.
For $\ord=1$, the circuit is shown on the left in \cref{fig:qn-1}; for $\ord=2$ on the right.
Higher orders can be constructed recursively.

When executed on pure states, this quantum neuron is a so-called repeat-until-success (RUS) circuit, meaning that the ancilla that is measured (purple meter in \cref{fig:qn-1}) indicates whether the circuit has been applied successfully.
When the outcome is zero, the neuron has been applied.
When the outcome is one, a (short) correction circuit reverts the state to its initial configuration.
Started from a pure state (e.g.\ $\ket x$ for $x\in \{0,1\}^n$, as above) and repeating whenever a $1$ is measured, one obtains an arbitrarily-high success probability.

Unfortunately this does not work for non-pure inputs.
For states in superposition, such as a state $(\ket x + \ket y)/\sqrt{2}$, for $x\neq y$ two bit-strings of length $n$, the amplitudes within the superposition will depend on the history of success \cite[appdx.~B]{Cao2017}.
Using a technique called fixed-point oblivious amplitude amplification \cite{Guerreschi2019,Tacchino2019}, one can alleviate this issue, and essentially post-select on measuring outcome $0$ while preserving unitarity of the operation to arbitarily high accuracy.
This comes at the cost of performing multiple rounds of these quantum circuits (and their inverses), the number of which will depend on the likelihood of measuring a zero (i.e.\ success) in first place.
This naturally depends on the parameters of the neuron, $\theta$, and the input state given.
In the following we will thus simply assume that the approximate postselection is possible, and carefully \emph{monitor} the overhead due to the necessary amplitude amplification in our empirical studies in \cref{sec:empirical}; we found that in general the postselection overhead remained mild, and tended to converge to zero as learning progressed.

The specific activation function this quantum neuron gives rise to is depicted in \cref{fig:activation}.
We point out that other shapes of activation functions can readily be implemented in a similar fashion \cite{DePaulaNeto2019}.

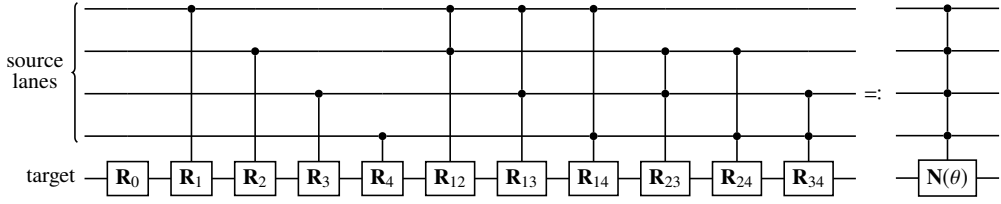
\begin{figure}[t]
    \centering
    \begin{tikzpicture}[scale=0.75]
    \node[scale=0.75] at (0, 0) {\begin{quantikz}[qc]
    \lstick[4]{\shortstack{source\\lanes}}
    & \qw & \ctrl4 & \qw   & \qw    & \qw    & \ctrl1 & \ctrl2 & \ctrl3 & \qw    & \qw    & \qw    & \qw \\
    & \qw & \qw   & \ctrl3 & \qw    & \qw    & \ctrl3 & \qw    & \qw    & \ctrl1 & \ctrl2 & \qw    & \qw \\
    & \qw & \qw   & \qw    & \ctrl2 & \qw    & \qw    & \ctrl2 & \qw    & \ctrl2 & \qw    & \ctrl1 & \qw \\
    & \qw & \qw   & \qw    & \qw    & \ctrl1 & \qw    & \qw    & \ctrl1 & \qw    & \ctrl1 & \ctrl1 & \qw \\
    \lstick{target} & \gate{\op R_0} & \gate{\op R_1} & \gate{\op R_2} & \gate{\op R_3} & \gate{\op R_4} & \gate{\op R_{12}} & \gate{\op R_{13}} & \gate{\op R_{14}} & \gate{\op R_{23}} & \gate{\op R_{24}} & \gate{\op R_{34}} & \qw
    \end{quantikz}$\eqqcolon$\begin{quantikz}[qc]
    \qw & \ctrl1 & \qw \\
    \qw & \ctrl1 & \qw \\
    \qw & \ctrl1 & \qw \\
    \qw & \ctrl1 & \qw \\
    \qw & \gate{\op N(\theta)} & \qw
    \end{quantikz}};
    \end{tikzpicture}
    \caption{Degree $d=2$ controlled rotation for quantum neuron shown in \cref{fig:qn-1}, on $n=4$ input neurons; the controlled rotations are $\op R_I \coloneqq \op R(\theta_I)$ for $I\subset[n]$ with $|I|\le d$. The quantum neuron thus carries a parameter vector $\theta \in \field R^D$ for $D=\sum_{i=0}^d\binom{n}{i}$.}
    \label{fig:neuron-rotation}
\end{figure}
In this work, we generalize this quantum neuron first proposed in \cite{Cao2017} by increasing the number of control terms.
More concretely, $\eta$ as given in \cref{eq:eta} is an affine transformation of the boolean vector $x=(x_1,\ldots,x_n)$ for $x_i \in \{0, 1\}$.
By including multi-control gates---with their own parametrized rotation, labelled by a multiindex $\theta_I$ depending on the qubits $i \in I$ that the gate is conditioned on---we obtain the possibility to include higher degree polynomials, namely
\begin{equation}\label{eq:eta'}
    \eta' = \theta_0 + \sum_{i=1}^n \theta_i x_i + \sum_{i=1}^n \sum_{j=1}^n \theta_{ij} x_i x_j + \ldots = \sum_{\substack{I \subseteq [n]\\|I| \le d}} \theta_I \prod_{i\in I} x_i,
\end{equation}
where $d$ labels the degree of the neuron; for $d=2$ and $n=4$ an example of a controlled rotation that gives rise to this higher order transformation $\eta'$ on the bit string $x_i$ is shown in \cref{fig:neuron-rotation}.
In this fashion, higher degree boolean logic operations can be directly encoded within a single conditional rotation: an AND operation between two bits $x_1$ and $x_2$ is simply $x_1x_2$.

\subsection{QRNN Cell}\label{sec:cell}
The quantum neuron defined in \cref{sec:qn} will be the crucial ingredient in the construction of our quantum recurrent neural network cell.
Much like for classical RNNs and LSTMs, we define such a cell which will be applied iteratively to the input presented to the network.
More specifically, the cell is comprised of in- and output lanes that are reset after each step, as well as an internal cell state which is passed on to the next iteration of the network.
The setup and inner workings of this cell are depicted and described in \cref{fig:qrnn-cell}.

\begin{figure}[t]
    \hspace{-2cm}
    \newcommand\ggrinput{\gategroup[wires=6,steps=5,style={dashed,inner xsep=2pt}]{input stage}}
    \newcommand\ggrstage{\gategroup[wires=6,steps=5,style={dashed,inner xsep=2pt}]{work stages $s=1,\ldots,S$}}
    \newcommand\ggroutput{\gategroup[wires=6,steps=5,style={dashed,inner xsep=2pt}]{output stage}}
    \newcommand\wiredots{\ \ldots\ \qw}
    \newcommand{\circuit}{\begin{quantikz}[qc]
	\lstick[wires=3]{\rotatebox{90}{cell state in}}  & \qw & \qw\ggrinput                & \gate{\op N_1^\mathrm{in}} & \qw                        & \wiredots & \qw                        & \qw & \gate{\op R_1^s}\ggrstage & \gate{\op N_1^s} & \ctrl1           & \wiredots & \ctrl 1          & \qw & \qw\ggroutput                                                        & \ctrl1                      & \ctrl1                      & \wiredots & \ctrl1                      & \qw \rstick[wires=3]{\rotatebox{90}{cell state out}} \\
	                                           & \qw & \qw                         & \qw                        & \gate{\op N_2^\mathrm{in}} & \wiredots & \qw                        & \qw & \gate{\op R_2^s}          & \ctrl{-1}        & \gate{\op N_2^s} & \wiredots & \ctrl1           & \qw & \qw                                                                  & \ctrl1                      & \ctrl1                      & \wiredots & \ctrl1                      & \qw                                            \\[.5cm]
	                                           & \qw & \qw                         & \qw                        & \qw                        & \wiredots & \gate{\op N_H^\mathrm{in}} & \qw & \gate{\op R_H^s}          & \ctrl{-1}        & \ctrl{-1}        & \wiredots & \gate{\op N_H^s} & \qw & \qw                                                                  & \ctrl1                      & \ctrl2                      & \wiredots & \ctrl3                      & \qw                                            \\
	\lstick[wires=3]{\rotatebox{90}{i/o lanes $\ket0^{\otimes I}$}} & \qw & \gate[3]{\op X}             & \ctrl{-3}                  & \ctrl{-2}                  & \wiredots & \ctrl{-1}                  & \qw & \qw                       & \ctrl{-1}        & \ctrl{-1}        & \wiredots & \ctrl{-1}        & \qw & \gate[wires=3]{\op X}                                                & \gate{\op N_1^\mathrm{out}} & \qw                         & \wiredots & \qw                         & \meter{}                                       \\
	                                           & \qw &                             & \ctrl{-1}                  & \ctrl{-1}                  & \wiredots & \ctrl{-1}                  & \qw & \qw                       & \ctrl{-1}        & \ctrl{-1}        & \wiredots & \ctrl{-1}        & \qw &                                                                      & \qw                         & \gate{\op N_2^\mathrm{out}} & \wiredots & \qw                         & \meter{}\vcw{-1}                               \\[.5cm]
	                                           & \qw &                             & \ctrl{-1}                  & \ctrl{-1}                  & \wiredots & \ctrl{-1}                  & \qw & \qw                       & \ctrl{-1}        & \ctrl{-1}        & \wiredots & \ctrl{-1}        & \qw &                                                                      & \qw                         & \qw                         & \wiredots & \gate{\op N_I^\mathrm{out}} & \meter{}\vcw{-1}                               \\[.5cm]
	                                           &     & \lstick{input word}\vcw{-1} &                            &                            &           &                            &     &                           &                  &                  &           &                  &     & \lstick{input word}\vcw{-1}\rstick{\raisebox{.7mm}{(resets i/o lanes)}} &                             &                             &           &                             & \vcw{-1} \rstick{\raisebox{.7mm}{\shortstack[l]{output\\word}}}
    \end{quantikz}}
    \begin{tikzpicture}[scale=0.9]
    \node[anchor=south west,scale=0.8] at (0, 0) {\circuit};
    \node at (1.2, 4.8) {$\vdots$};
    \node at (1.2, 2.35) {$\vdots$};
    \end{tikzpicture}
    \caption{Quantum recurrent neural network cell.
    Each controlled quantum neuron $\op{cN}^i_j$ is implemented as explained in \cref{sec:qn,fig:qn-1} and comes with its own parameter vector $\theta^i_j$, where we draw the control lanes from the rotation inputs as depicted in \cref{fig:neuron-rotation}, with ancillas omitted for clarity.
    The $\op R^i_j$ are extra rotations with a separate parameter set $\phi^i_j$.}
    \label{fig:qrnn-cell}
\end{figure}
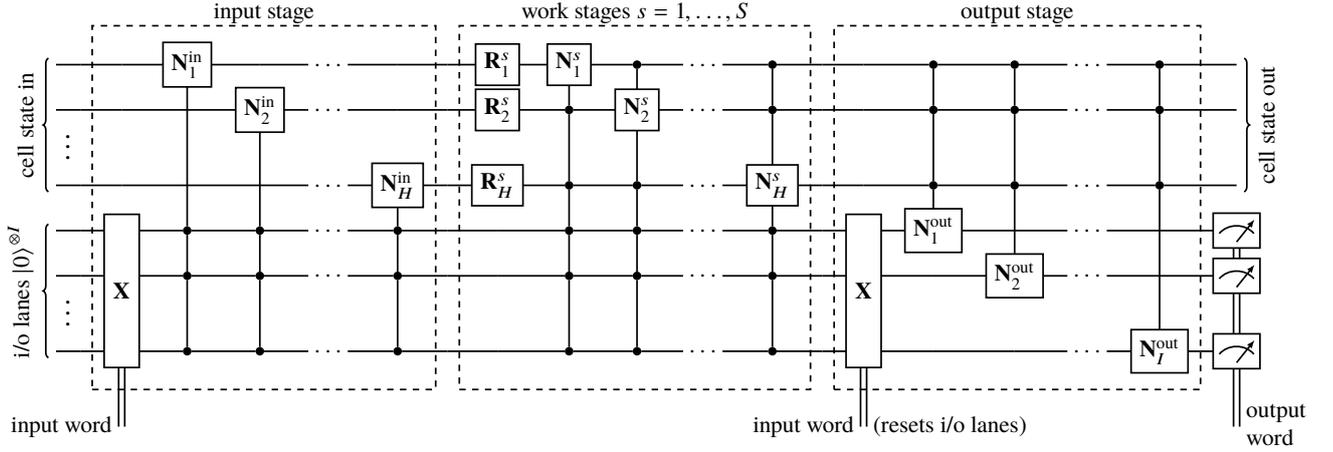

\begin{figure}[t]
    \hspace{-0.7cm}
    \newcommand\ggrcell{\gategroup[wires=3,steps=8,style={dashed,inner xsep=2pt}]{cell}}    \newcommand\ggrcellfull{\gategroup[wires=3,steps=3,style={dashed,inner xsep=2pt,fill=white}]{cell}}
    \newcommand\wiredots{\ \ldots\ \qw}
    \newcommand\cwiredots{\ \ldots\ \cw}
    \newcommand{\circuit}{
    \begin{quantikz}[qc,qctight]
	\lstick{$\ket0$} & \qwbundle{H}\qw & \qw & \ggrcell\qw                           & \gate{\op N} & \gate[2]{\rotatebox{90}{\makebox[5mm]{stage 1}}} & \wiredots  & \gate[2]{\rotatebox{90}{\makebox[5mm]{stage s}}} & \qw          & \ctrl1       & \qw                                   & \qw & \qw & \qw & \qw & \qw              & \qw & \ggrcellfull\qw &                                      &                                       & \qw & \qw & \wiredots & \qw & \qw              & \qw & \ggrcellfull\qw &                                      &                                       & \qw & \qw\\
	\lstick{$\ket0$} & \qwbundle{I}\qw & \qw & \gate{\op X}                          & \ctrl{-1}    &                                                  & \wiredots  &                                                  & \gate{\op X} & \gate{\op N} & \meter{}                              &     &     &     &     & \lstick{$\ket0$} & \qw & \qw             & \gate{\op X}                         & \gate{\op X}                          &     &     &           &     & \lstick{$\ket0$} & \qw & \qw             & \gate{\op X}                         & \gate{\op X}                          &     \\
	                 &                 &     & \cctrl{-1}                            & \cw          & \cw                                              & \cwiredots & \cw                                              & \cwbend{-1}  &              &                                       &     &     &     &     &                  &     &                 &                                      &                                       &     &     &           &     &                  &     &                 &                                      &                                       &     \\
	                 &                 &     & \vcw{-1} \rstick{in$_1$} &              &                                                  &            &                                                  &              &              & \vcw{-2}\rstick{out$_1$} &     &     &     &     &                  &     &                 & \vcw{-2}\rstick{in$_2$} & \vcw{-2}\rstick{out$_2$} &     &     &           &     &                  &     &                 & \vcw{-2}\rstick{in$_L$} & \vcw{-2}\rstick{out$_L$} &
    \end{quantikz}}
    \begin{tikzpicture}
    \node[scale=0.88] at (0, 0) {\circuit};
    \end{tikzpicture}
    \caption{Quantum recurrent neural network, by applying the same QRNN cell constructed in \cref{sec:cell} iteratively to a sequence of input words $\text{in}_1, \ldots, \text{in}_L$.
    All input and ancilla qubits used throughout can be reused; we thus need $H+I+\ord$ qubits, where $H$ is the cell state workspace size, $I$ the input token width (in bits), and $\ord$ the order of the quantum neuron activation, as explained in \cref{sec:qn}.
    }
    \label{fig:qrnn}
\end{figure}
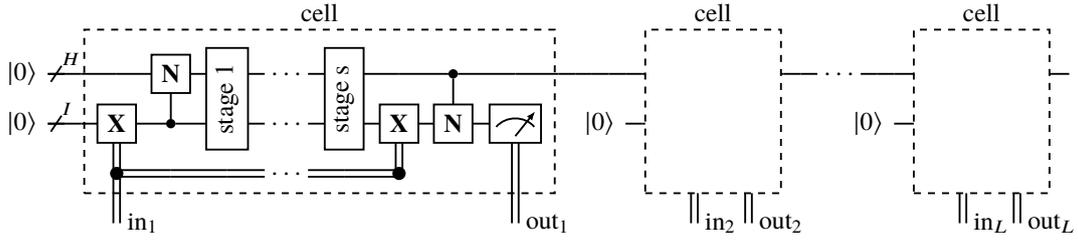

\subsection{Sequence to Sequence Model}\label{sec:model}

In order to be able to apply the QRNN cell constructed in \cref{sec:cell}, we need to iteratively apply it to a sequence of input words $ \text{in}_1, \text{in}_2, \ldots, \text{in}_L$.
This is achieved as depicted in \cref{fig:qrnn}.

The output lanes $\mathrm{out}_i$ label a measured discrete distribution $p_i$ over the class labels (which we can do by reading out the statevector weights if running a simulation on a classical computer; or by repeated measurements on quantum hardware).
This distribution can then be fed into an associated loss function such as cross entropy or CTC loss.

\section{Implementation and Training}
We implemented the QRNN in pytorch \cite{code}, using custom quantum gate layers and operations that allow us to extract the predicted distributions at each step.
As this is in essence a simulation of a quantum computation, we take the following shortcuts: instead of truly performing fixed-point amplitude amplification for the quantum neurons and output lanes during training, we postselect; as aforementioned, we kept track of the postselection probabilities which allows us to compute the overhead that would be necessary for amplitude amplification.
We further extract the output probability distribution instead of estimating it at every step using measurements.

In our experiments we focus on character-level RNNs.
Just like in the classical case, the sequence of predicted distributions $\{ p_i \}$ is fed into a standard \texttt{nn.CrossEntropyLoss} to minimize the distance to a target sequence.
With pytorch's autograd framework we are able to perform gradient-based learning directly; on quantum hardware either gradient-free optimizers such as L-BFGS, NatGrad would have to be utilized, or numerical gradients extracted \cite{Wierichs2020}.
All experiments were executed on 2-8 CPUs, and required between 500MB and 35GB of memory per core.
We implement a mechanism train batches of data in parallel.
A straightforward implementation in pytorch, on real-world hardware batching would simply mean executing the QRNN with the same parameters on multiple devices in parallel, and averaging the resulting losses.


\begin{figure}[t]
    \centering
    \includegraphics[width=\textwidth]{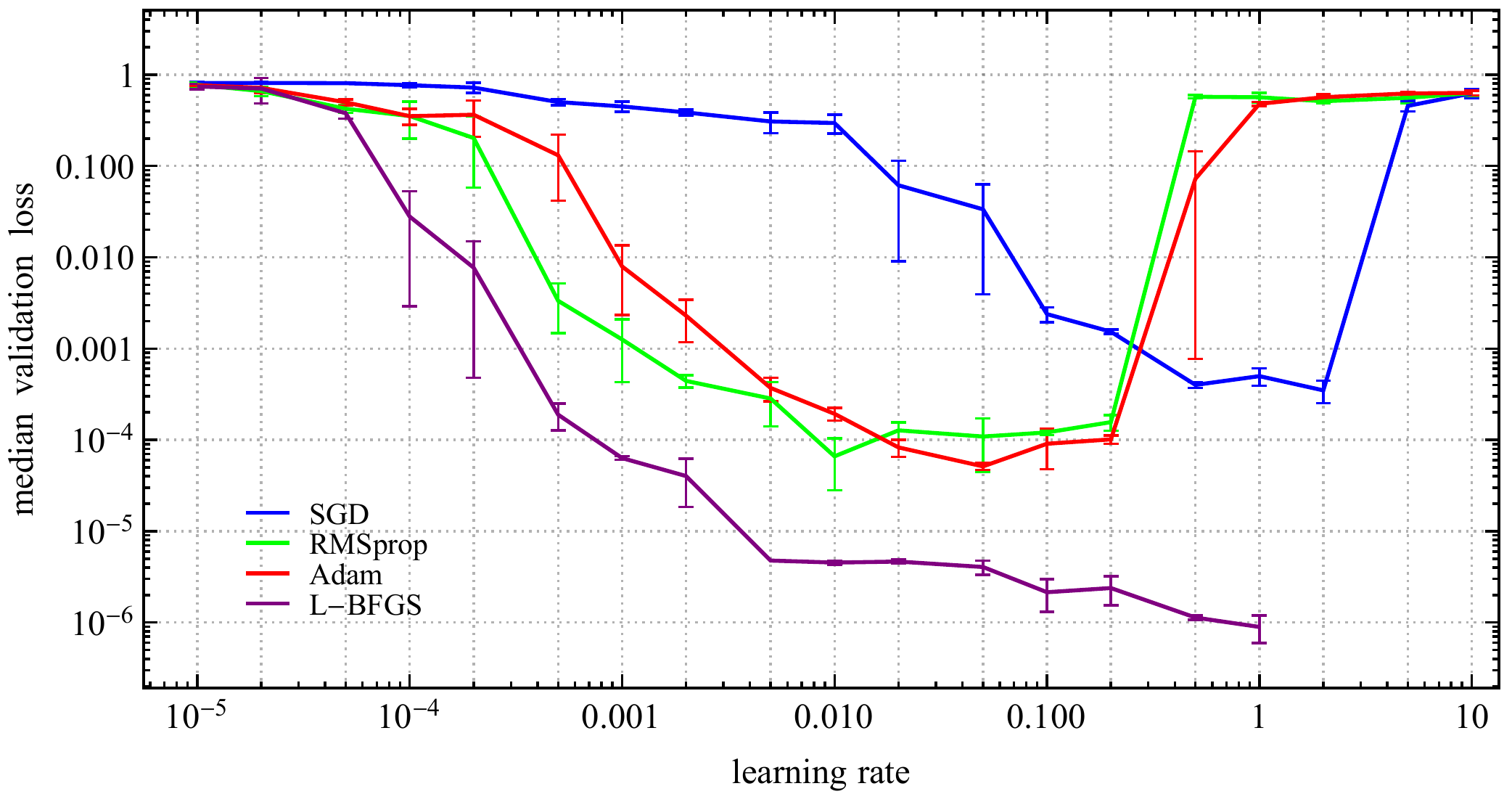}
    \caption{Training performance of SGD, RMSprop, Adam and L-BFGS optimizers over a range of learning rates, as described in \cref{sec:memo}.
    Shown is the validation loss achieved after 500 training steps.
    The plotted points indicate the median of five runs; the error bars indicate the median absolute deviation, a robust measure of dispersion within a univariate dataset.}
    \label{fig:optimizer}
\end{figure}

\section{Empirical Results}\label{sec:empirical}
\subsection{Sequence Memorization}\label{sec:memo}
The first task we implement is whether the network can learn to reproduce the two sequences \texttt{44444$\ldots$4} and \texttt{12312$\ldots$3}.
For a QRNN with 2 stages, neuron degree 3 and a workspace size of 5 (1162 parameters) this poses no problem; in fact, this is vastly over-parametrized for the task at hand, and good convergence can be achieved with a much smaller network.
Nonetheless, with this setup we benchmark optimizer and learning rate hyperparameters; our findings are summarized in \cref{fig:optimizer}.

While it tended to produce the lowest validation loss over a wide range of learning rates, we found the L-BFGS optimizer commonly used with VQE circuits to behave unpredictably for all but the simple sequence learning tasks: for a large set of initial seeds, only very few resulted in a good convergence within 500 training iterations---a problem that Adam did not exhibit, where generally many random initializations resulted in a good training run.
Furthermore, L-BFGS is extremely costly, taking about an order of magnitude longer for convergence and a significantly higher memory consumption than the other, purely gradient-based methods.

SGD has a narrow window of good learning rates, whereas RMSprop and Adam are less sensitive to this choice, with Adam generally outperforming the former in convergence time.
Due to its performance and relative robustness with respect to the choice of learning rate, Adam was our default choice for all following experiments if not stated otherwise; learning rates were generally kept at or around $0.05$.

In \cref{fig:postselection} we plot the postselection overhead during training.
Notably, the overhead quickly converges to a factor of one as the validation loss decreases.

\subsection{Finding Structure in Time}\label{sec:elman}
In his \citeyear{Elman1990} paper, \citeauthor{Elman1990} describes two basic sequence learning tasks to evaluate structure in time \cite{Elman1990}.
The first task is that of learning XOR sequences, which are binary strings $s=s_1 s_2 s_3 \ldots s_L$ such that each third digit is the XOR value of the preceding two, i.e.\ $s_{3i} = s_{3i-1} \oplus s_{3i-2}$; one example being $s=\texttt{000\,011\,110\,011\,101}$.

Due to the simplicity of the test, we use a QRNN with workspace size 4 and a single work stage to explore which parameter initialization converges to a validation loss threshold of $10^{-3}$ first.
As shown in \cref{fig:qrnn-cell}, there are two groups of parameters:
those for the neurons, and those for the single-qubit unitaries within the work stages.
Each quantum neuron as depicted in \cref{fig:neuron-rotation,eq:eta'} itself comprises two parameter sets: a bias gate $\op R_0$ with angle $\theta_0$, and the weights (all other parameters). Choosing to initialize each of them with a normal distribution with mean $\mu$ and width $\sigma$, we have four parameter group hyperparameters: bias $\mu$, bias $\sigma$, weights $\sigma$ (the mean is already captured in the bias gate), and unitaries $\sigma$ (for which we chose the mean to be zero by default).

Our findings and choices for the default initialisation are collected in \cref{fig:initialization} in the appendix.
The most influential meta parameter is the bias $\mu=\pi/4$---which, as shown in \cref{fig:activation}, places the initial polynomial $\eta$ at the steepest slope of the activation function, which results in a large initial gradient.

The second task described in \cite{Elman1990} is that of learning the structure of a sentence made up of the three words ``ba'', ``dii'' and ``guuu''; an example being ``ba\,dii\,ba\,guuu\,dii''.
Having seen the letter `d', the network thus has to remember that the next two letters to predict are ``ii'', and so on.
We chose this slightly more difficult task to assess how the QRNN topology influence convergence speed.
We found that the neuron order $\ord=2$ performs best, which results in an activation function with relatively steep flanks and flat plateau sections.
The latter allow the incoming signal to remain constant for a significant range over its parameters, while still maintaining a small but nonzero gradient throughout the plateau area (this plateau gradient is significantly suppressed already at $\ord=3$).
The other parameters are discussed in \cref{sec:apdx-params}.

\subsection{MNIST Classification}\label{sec:mnist}
\begin{table}[]
    \hspace*{-.5cm}
    \begin{tabular}{r llll}
    \toprule
        Digit Set & Method & Data Augmentation & Accuracy $[\%]$ \\
    \midrule
        \multirow{2}{*}{$\{0, 1\}$} & QRNN (12 qubits, Adam) & none & $99.2\pm0.2$ \\
        & \emph{\ \ ensemble of 4} & none & $\mathbf{99.6\pm0.16}$ \\
    \midrule
        \multirow{3}{*}{$\{3, 6\}$} & VQE (17 qubits) \cite{Farhi2018}$^{\mathsection}$ & ambiguous samples removed & 98 
        \\
        & QRNN (12 qubits, Adam) & none & $89.7\pm 0.8$\\
        & QRNN (10 qubits, L-BFGS) & none & $97.1\pm0.7$\\
        & \emph{\ \ ensemble of 6} & none & $\mathbf{99.0\pm0.3}$ \\
    \midrule
        \multirow{7}{*}{full MNIST} & VQE (10 qubits) \cite{Grant2019}$^{\dagger}$ & partitioned into $\{\text{even}, \text{odd}\}$ & 82 \\
        & uRNN \cite{Arjovsky2015} & none & 95.1 \\
        & LSTM \cite{Arjovsky2015} & none & 98.2 \\
        & QFD ($>200$ qubits) \cite{Kerenidis2018}$^\ddagger$ & PCA, slow feature analysis & 98.5\\
        & QRNN (10 qubits, Adam) & PCA, t-SNE & $94.6\pm0.4$ \\
        & QRNN (13 qubits, Adam) & UMAP & $96.7\pm0.2$ \\
        & \emph{\ \ ensemble of 3} & UMAP & $\mathbf{99.23\pm0.05}$ \\
    \bottomrule
    \end{tabular}
    \vspace*{2mm}
    \caption{Classification of MNIST using QRNNs on 12 qubits (workspace size 8, 2 stages, neuron degree 2; 1212 parameters), 10 qubits (workspace size 6, 2 stages, neuron degree 3; 1292 parameters), and 13 qubits (workspace size 7, 2 stages, neuron degree 3; 3134 parameters), as well as ensembles thereof.
    Input either presented pixel-by-pixel; resp.\ t-SNE/UMAP augmented with discretized coordinates ($2$ to $4$ dimensions, presented bit by bit up to 8 bits of precision).\\
    \textsuperscript{$\mathsection$})~Images down-scaled to $4\times4$ pixels and binarized. The set used for the accuracy measurement thus contains $\approx 70\%$ samples already present during training. Naturally, this is a problem of the small sample dimensions; and to varying extent also present in the full MNIST classification models below that employ other dimensionality reduction techniques. Our $10\times10$ pixel-by-pixel data contains 80 duplicates for the digit \texttt{`1`} between training and test set (so $\approx 7\%$ of the test class), and a single duplicate for the digit \texttt{`7`} (so $\approx 0.1\%$ of the test class). \\
    \textsuperscript{$\dagger$})~Image data prepared in superposition, i.e.\ as a state $\propto \sum_{i}v_i\ket i=(v_1, \ldots, v_n)$, where $v_i\in[0, 1]$ is the pixel value at position $i$ in the image; VQE with such an embedding thus serve as linear discriminator.\\
    \textsuperscript{$\ddagger$})~The paper exploits a well-known quantum speedup in performing sparse linear algebra \cite{Harrow2009a} to implement a quantum variant of slow feature analysis as performed in \cite{berkes2005pattern}, but with a different classifier. It relies on quantum random access memory (QRAM), recently shown to allow ``de-quantization'' of claimed exponential quantum speedups \cite{Tang2019}. The authors do not explicitly state a qubit count, so it is lower-bounded from the number of qubits required for the QRAM alone.}
    \label{tab:mnist}
\end{table}

To test whether we can utilize our QRNN setup to classify more complex examples, we assess its performance on handwritten integer digit classification.
While this is a rather untypical task for a recurrent network, there exist baselines both for a comparison with a classical RNN, as well as with quantum classifiers.

We use the MNIST dataset, with a $55010:5000:10000$ train~:~validate~:~test split, where the training subset is the standard one provided with the dataset, and the validate~:~test split is chosen manually at random from the provided test set \cite{lecun2010mnist}.
As a post-processing step we first crop the images to $20\times 20$ pixels, downscale them to size $10\times10$, and then binarize each image such that it has a bit depth of 1.
Naturally, this significantly reduces the information content of the training data; yet as we are not competing with top-of-the-line classifiers with our QRNN model we found the resulting input sequence length of 100 (for the $10\times10$ pixel images) to give a good compromise between computational cost, classification performance, and visual acuity---the latter is of particular importance for the generative task, where we wish to qualitatively assess that the network can render handwritten digits.

We choose two ``scanlines'' across each image: one left-to-right, top-to-bottom; the other one top-to-bottom, left-to-right.
This means that two bits of data are presented at each step.
The output labels are then simply binary values of the numbers $0$ to $9$, which have to be written to the output at the last few steps of the sequence (in little Endian order).
We found that pairs of digits such as \texttt{`0'} and \texttt{`1'} (arguably the easiest ones) could be discriminated with $\approx 99.2\%$ success probability when using a single network ($\approx 99.6\%$ for an ensemble of four); but even more complicated pairs like `3' and `6' could be distinguished with $\approx 99\%$ likelihood.

\begin{figure}[t]
    \hspace*{-2cm}\begin{minipage}{18cm}\vspace*{-1cm}
    \includegraphics[width=18cm]{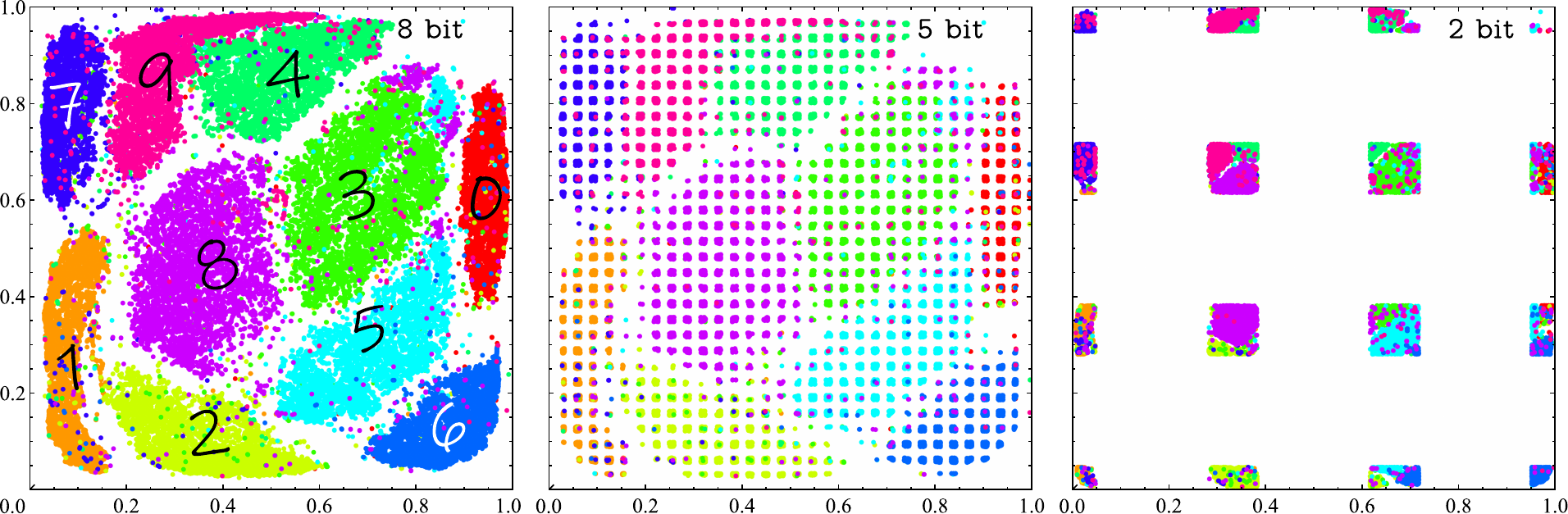}
    \end{minipage}
    \caption{MNIST training dataset mapped to a 2D plane by t-SNE; where each coordinate is further mapped by an inverted Gaussian to distribute all samples approximately uniformly over the intervals $[0,1]$ in $x$ and $y$ direction, respectively.
    The left, middle and right pictures show discretized coordinates of 8, 5 and 2 bits, respectively.
    t-SNE is an unsupervised feature extractor; meaning the clusters of digits emerge without knowledge of the digits' labels.
    It is clear that the discretization step introduces additional errors; the lower the resolution of the coordinate, the more the information content is diminished.}
    \label{fig:mnist}
\end{figure}

In addition to classifying digits by presenting images pixel-by-pixel, we also used 2D and 3D t-distributed stochastic neighbor embedding (t-SNE, \cite{tSNE}) clustering as data augmentation in a first step; and a more modern dimensionality reduction technique called uniform manifold approximation and projection (UMAP, \cite{UMAP}), which reduces the 792-dimensional MNIST vectors to $2$ to $4$ dimensions.
The RNN was then presented with the up to four coordinates, discretized to between two and eight bits of precision, which the network has learn to decode and classify.
To aid in the resulting information loss due to the discretization of floating point values, we mapped the raw dimensionality-reduced coordinates to the discretized ones in a nonlinear fashion (an inverted Gaussian) to make full use of the available dynamic range.

It is worth emphasizing that similar to other feature extraction methods such as principal component analysis (PCA), all of the feature maps are learned \emph{only} on the training set, and then applied as-is to the validation and test sets.
This ensures that only information from the training samples is used in the creation of a preprocessing pipeline.
While t-SNE is extremely powerful at revealing clusters in data (see \cref{fig:mnist}), using it as a generic preprocessing step is computationally costly.
We can still employ it for a dataset the size of MNIST, by first performing PCA to reduce each image to 30 dimensions (instead of 100), and then---as a second step---t-SNE. This is a common chain of reductions and was e.g.\ also used in \cite{Kerenidis2018}, with a combination of PCA and slow feature analysis.

Perhaps unsurprisingly we found UMAP to outperform t-SNE by a large margin; this is to be expected, as UMAP is designed for dimensionality reduction, whereas t-SNE's primary goal is visualization of high-dimensional datasets.

We summarize our findings and a comparison with existing literature in \cref{tab:mnist}.

\subsection{QRNNs as Generative Models}\label{sec:gen}
\begin{figure}[t]
    \centering
    \includegraphics[width=10cm]{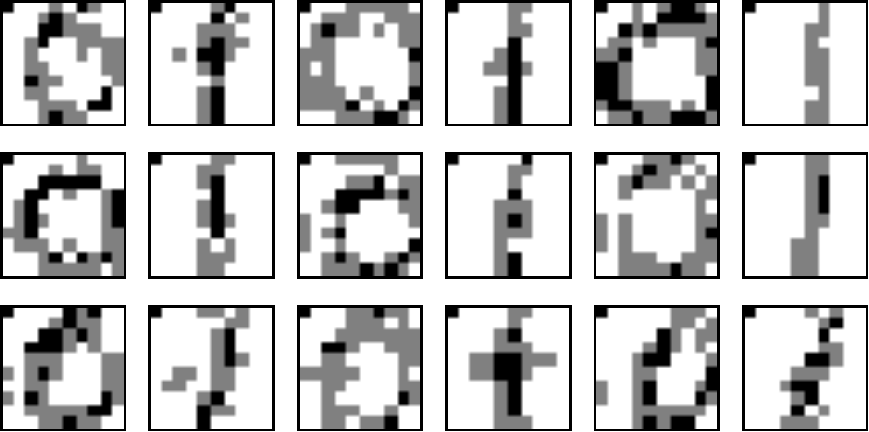}
    \caption{QRNN-generated handwritten digits \texttt{`0'} and \texttt{`1'}, as explained in \cref{sec:gen}.
    The QRNN topology is the same as for MNIST classification, with 1212 free parameters (\cref{sec:mnist}).}
    \label{fig:gen-digits}
\end{figure}

Instead of classifying digits, QRNNs can also be used as generative models.
With a \texttt{`0'} or \texttt{`1'} presented to the QRNN at the first step, we train it to re-generate handwritten digits.
As shown in \cref{fig:gen-digits}, the network indeed learns the global structure of the two digits; where it is evident that the intrinsic randomness due to quantum measurements present during inference gives rise to digits with various characteristica, such as character slant or line thickness.

\subsection{Long Sequence Tests}\label{sec:dna}
To assess our claims of high-quality gradients even in a situation of long sequences, we set up a test set consisting of gene sequences made up of the bases \texttt{`G'}, \texttt{`A'}, \texttt{`T'} and \texttt{'C'}; a single \texttt{`U'} is then inserted at a random position within the first half of the string, and the task of the network is to identify the base following after the \texttt{`U'}.
For instance, the label to identify for the sequence \texttt{`AGAUATTCAGAAT'} is \texttt{`A'}.
We repeat this classification task for multiple sequence lengths and several initial seeds, and stop after the validation loss is below a threshold of $10^{-3}$.
The number of steps required to reach this threshold---where at every training step the network is presented with a batch of 128 random training samples---is then our metric of success; the lower the better.

\begin{figure}[t]
    \centering
    \includegraphics[width=\textwidth]{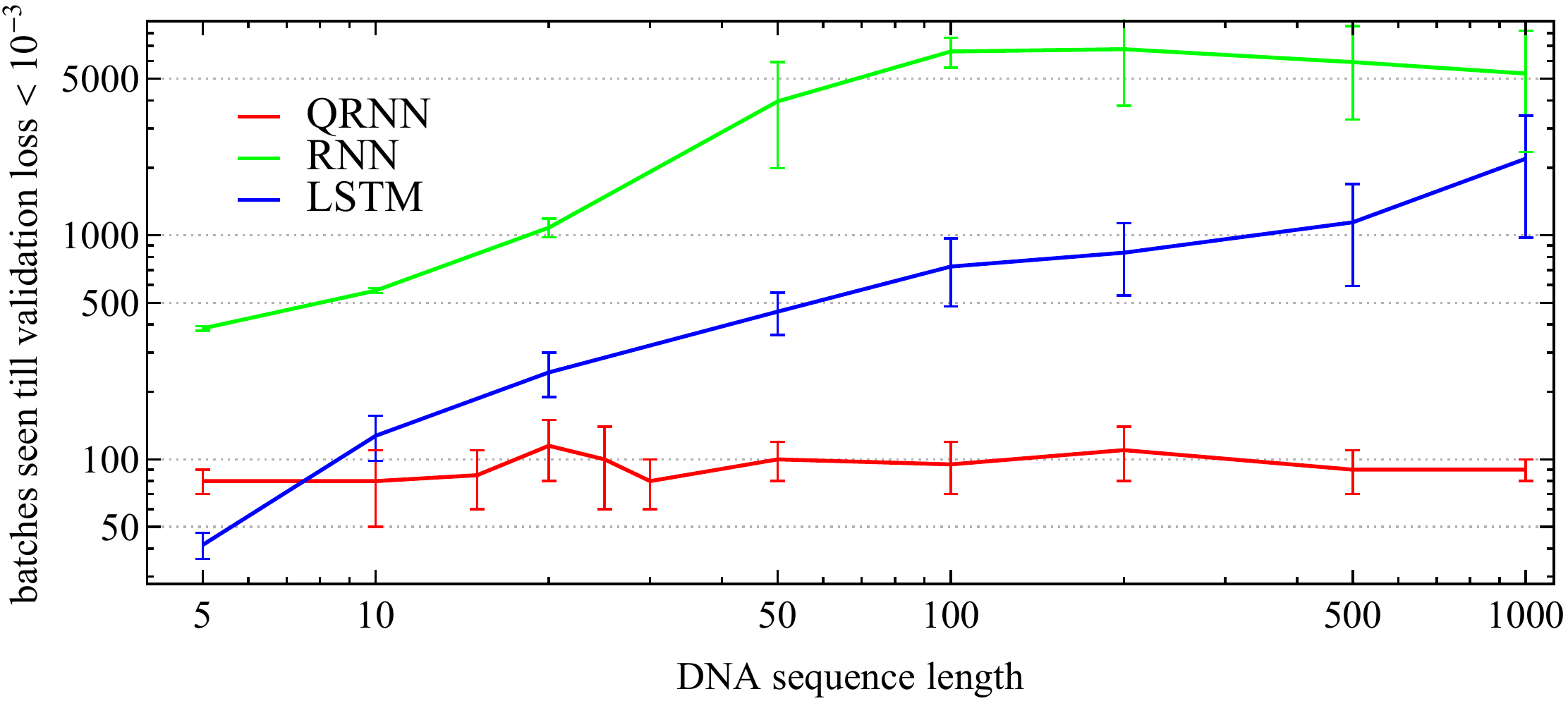}
    \caption{Median number of training steps to reach a validation loss $10^{-3}$ (with error bars indicating median absolute deviation).
    Shown are QRNN (red; 1 work stage, cell state size 5, 837 parameters), RNN (blue; 1 layers, hidden layer size 22, and a final linear layer; $888$ parameters) and LSTM (green; 1 layers, hidden layer size 10, and a final linear layer; $888$ parameters) for DNA sequence recognition task described in \cref{sec:dna}.}
    \label{fig:dna-times}
\end{figure}

We found that a QRNN with a workspace size of 5, one work stage, and activation degree 3 (resulting in 837 parameters) can be trained within an almost constant number of steps, even for string lengths of 1000 bases.

As comparison, we train an RNN and an LSTM variant on precisely the same dataset.
To field an objective view from the optimizer's perspective, we recreated topologies with a parameter count that matched the QRNN's 837 as closely as possible: both the RNN with one layer of width 22, and the LSTM with one layer of width 10 have 888 trainable parameters.
For each classical network we optimized the learning rate as hyperparameter, but left the other stock choices provided by pytorch.\footnote{Without doubt it will be possible to improve upon the convergence times for the task at hand with a more thorough analysis of hyperparameters, optimizers, and activation functions, or by allowing the network a larger capacity in terms of trainable parameters. We expect this to hold true for both traditional recurrent models, as well as the QRNN.}

The results of this test can be seen in \cref{fig:dna-times}.
While the QRNN features a relatively stable number of necessary training steps till convergence ($\approx 100$, which means it has seen $\approx 1.28 \times 10^4$ samples), the LSTM shows a steady increase---starting out faster than the QRNN, but ending up with a more than order-of-magnitude worse performance in this test.
The RNN was hardest to train for this task.\footnote{We note that for sequences of length $100$ or longer, several of the RNN runs timed out at $100k$ training steps, which is when we interrupted and re-started the training.
For RNN's, data above this point is thus to be taken with a grain of salt.}

\section{Conclusion and Outlook}
Without doubt, existing recurrent models---even simple RNNs---outclass the proposed QRNN architecture in this paper in real-world learning tasks.
In part, this is because we cannot easily simulate a large number of qubits on classical hardware: the memory requirements necessarily grow exponentially in the size of the workspace, for instance, which limits the number of parameters we can introduce in our model---on a quantum computer this overhead would vanish, resulting in a linear execution time in the circuit depth.

What should nevertheless come as a surprise is that the model \emph{does} perform relatively well on non-trivial tasks such as the ones presented here, in particular given the small number of qubits (usually between 8 and 14) that we utilised.
As qubit counts in real-world devices are severely limited---and likely will be for the foreseeable future---learning algorithms with tame system requirements will certainly hold an advantage.

Moreover, while we motivate the topology of the presented QRNN cell given in \cref{fig:qrnn-cell} by the action of its different stages (writing the input; work; writing the output), and while the resulting circuits are already far more structured than existing VQE setups, our architecture is still simplistic as compared to the various components of an RNN, let alone an LSTM.
In all likelihood, a more specialized circuit structure will outperform the ``simple'' quantum recurrent network presented herein.

Beyond the exploratory aspect of our work, our main insights are twofold.
On the classical side---as discussed in the introduction---we present an architecture which can run on current hardware and ML implementations such as pytorch; and which is a candidate parametrization for unitary recurrent models that hold promise in circumventing gradient degradation for long sequence lengths.
On the quantum side, we present a quantum machine learning model that allows ingestion of data of far more than a few bits of size; demonstrate that models with large parameter counts can indeed be evaluated and trained; and that classical baselines such as MNIST classification are within reach for variational quantum algorithms.

Variants of this ``quantum-first'' recurrent model might find application in conjunction with other quantum machine learning algorithms, such as quantum beam search \cite{Bausch2019c}, which could be employed in the context of language modelling.
With a more near-term focus in mind, modelling the evolution of quantum systems with noisy dynamics is a task that can be addressed using classical recurrent models \cite{Flurin2020}.
Due to the intrinsic capability of a QRNN to keep track of a quantum state, a quantum recurrent model might promise to better capture the exponentially-growing phase space dimension of the system under study.

\subsection*{Acknowledgements}
J.\,B.~acknowledges support of the Draper's Research Fellowship at Pembroke College, and wishes to thank Stephanie Hyland and Jean Maillard for helpful discussions and suggestions.

\printbibliography

\clearpage
\appendix

\begin{figure}[t]
    \thispagestyle{empty}
    \hspace*{-2.2cm}%
    \begin{minipage}{18cm}
    \centering\vspace*{-1.2cm}
    \includegraphics[width=.33\textwidth]{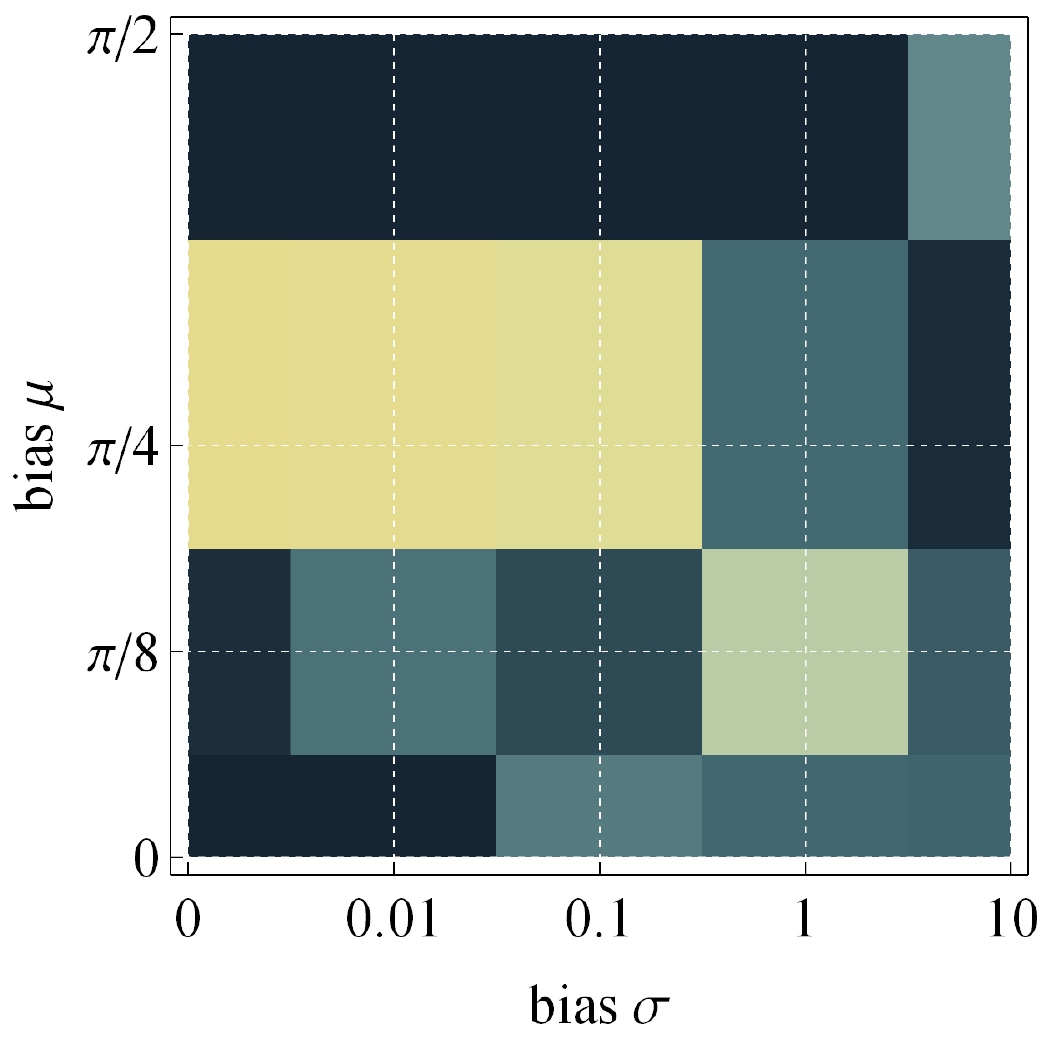}
    \includegraphics[width=.33\textwidth]{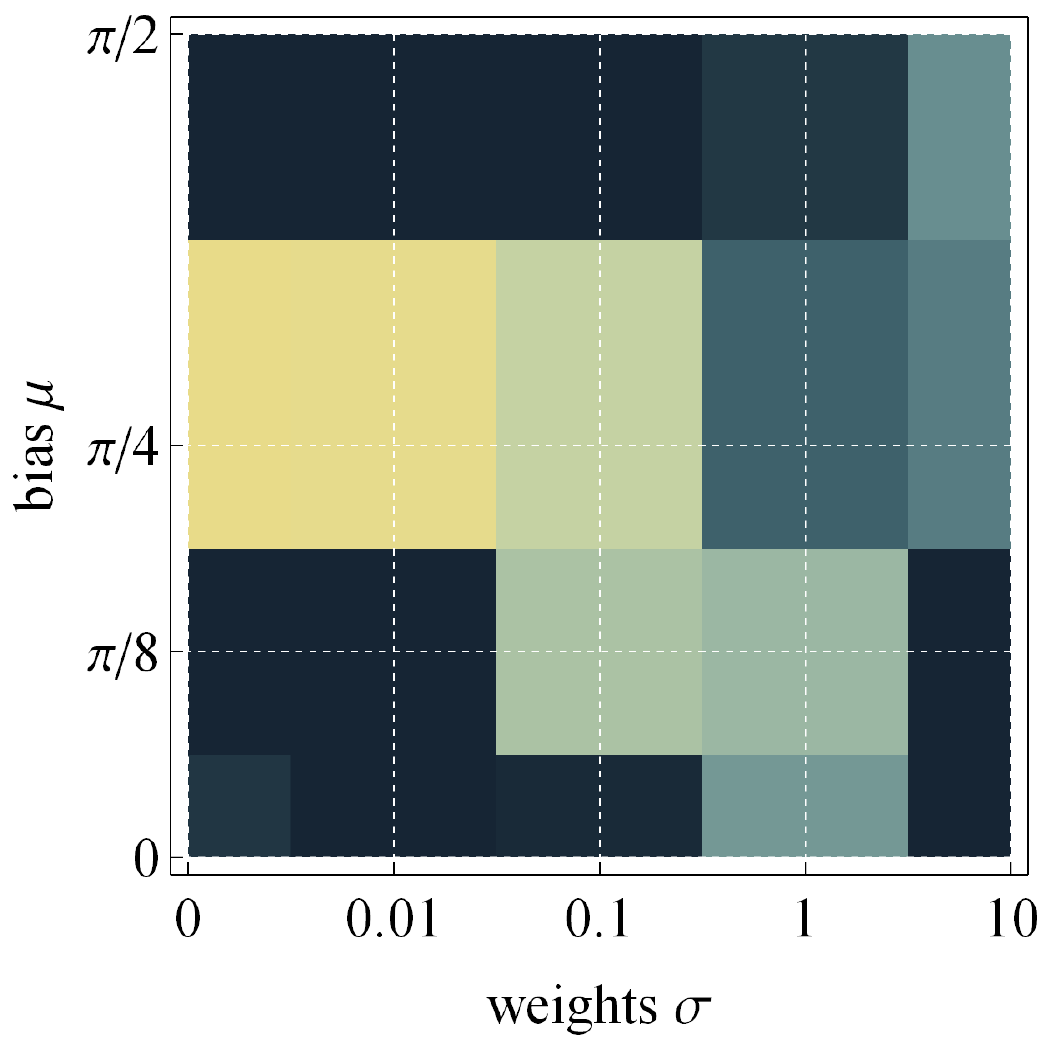}
    \includegraphics[width=.33\textwidth]{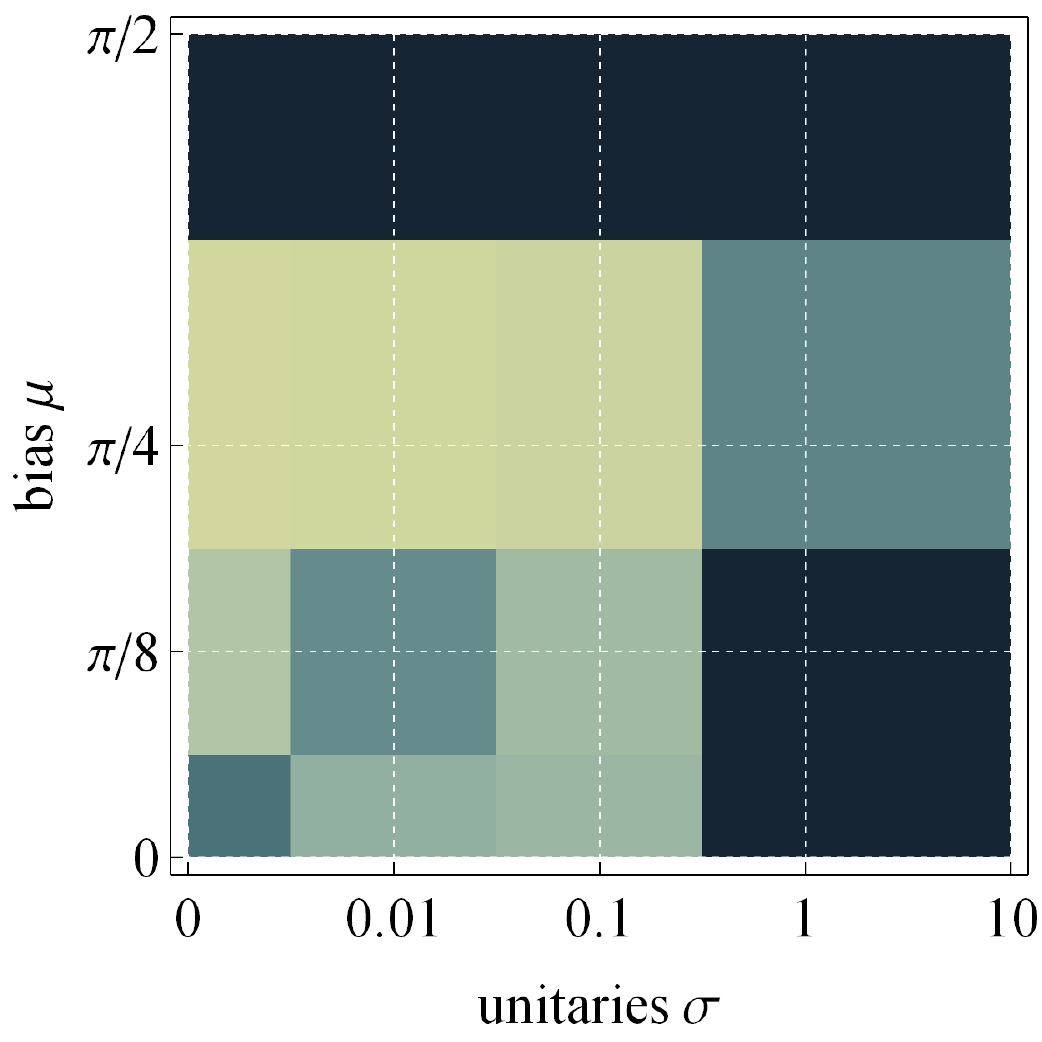}\\
    \includegraphics[width=.33\textwidth]{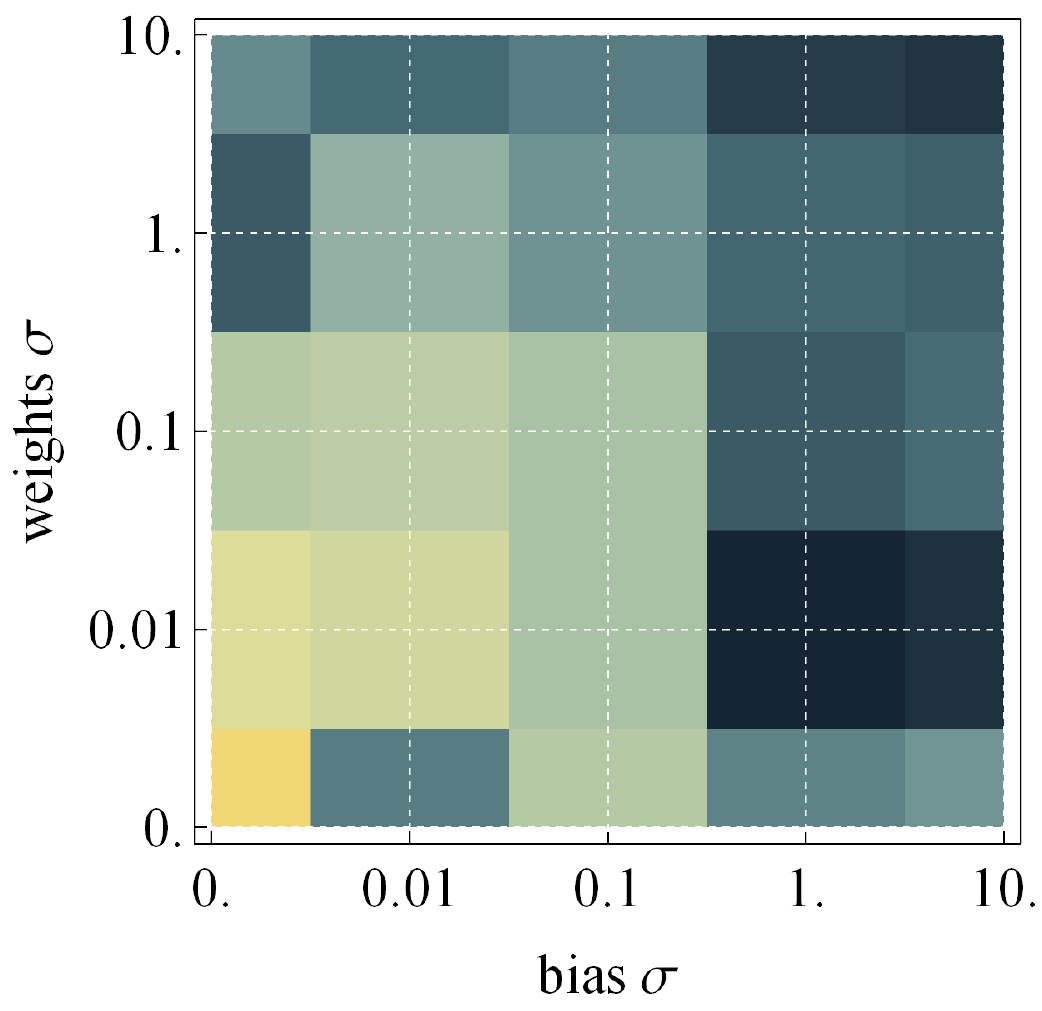}
    \includegraphics[width=.33\textwidth]{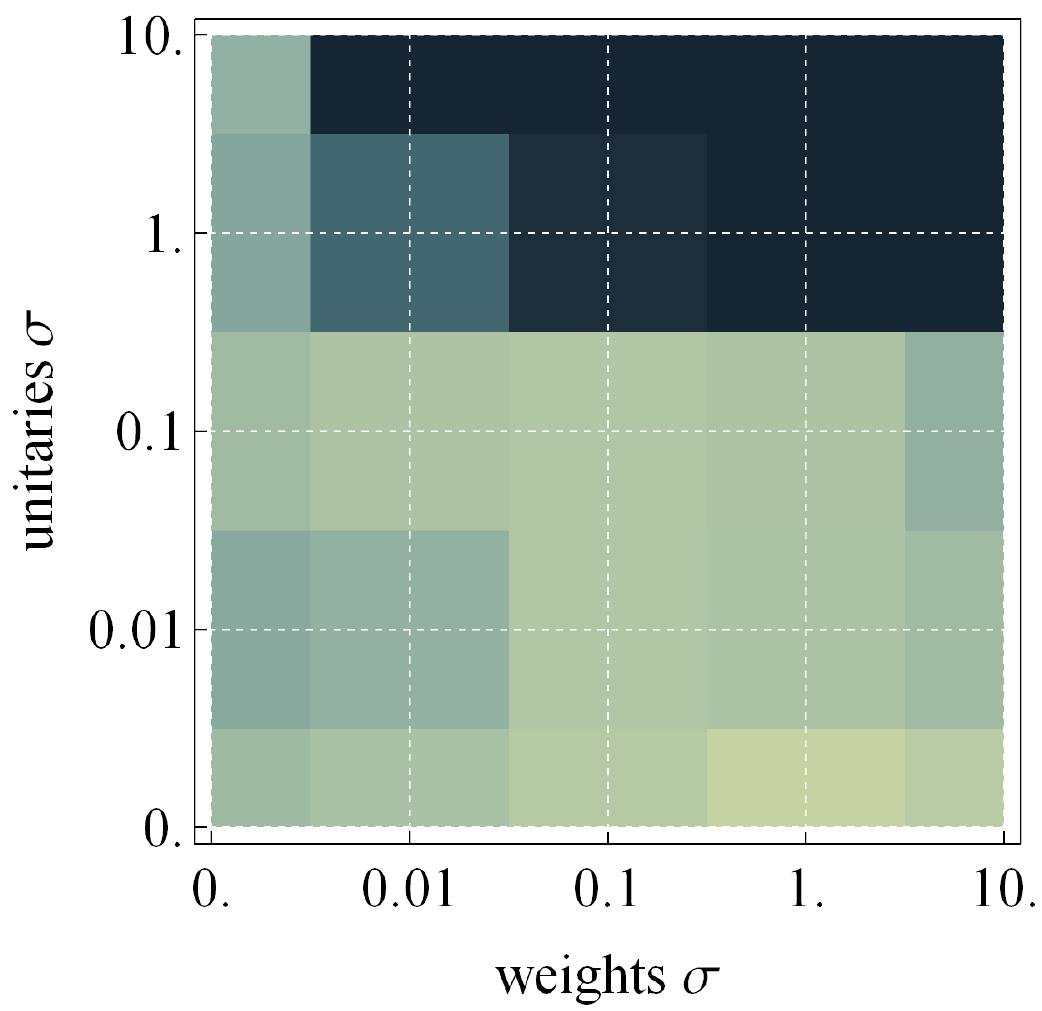}
    \includegraphics[width=.33\textwidth]{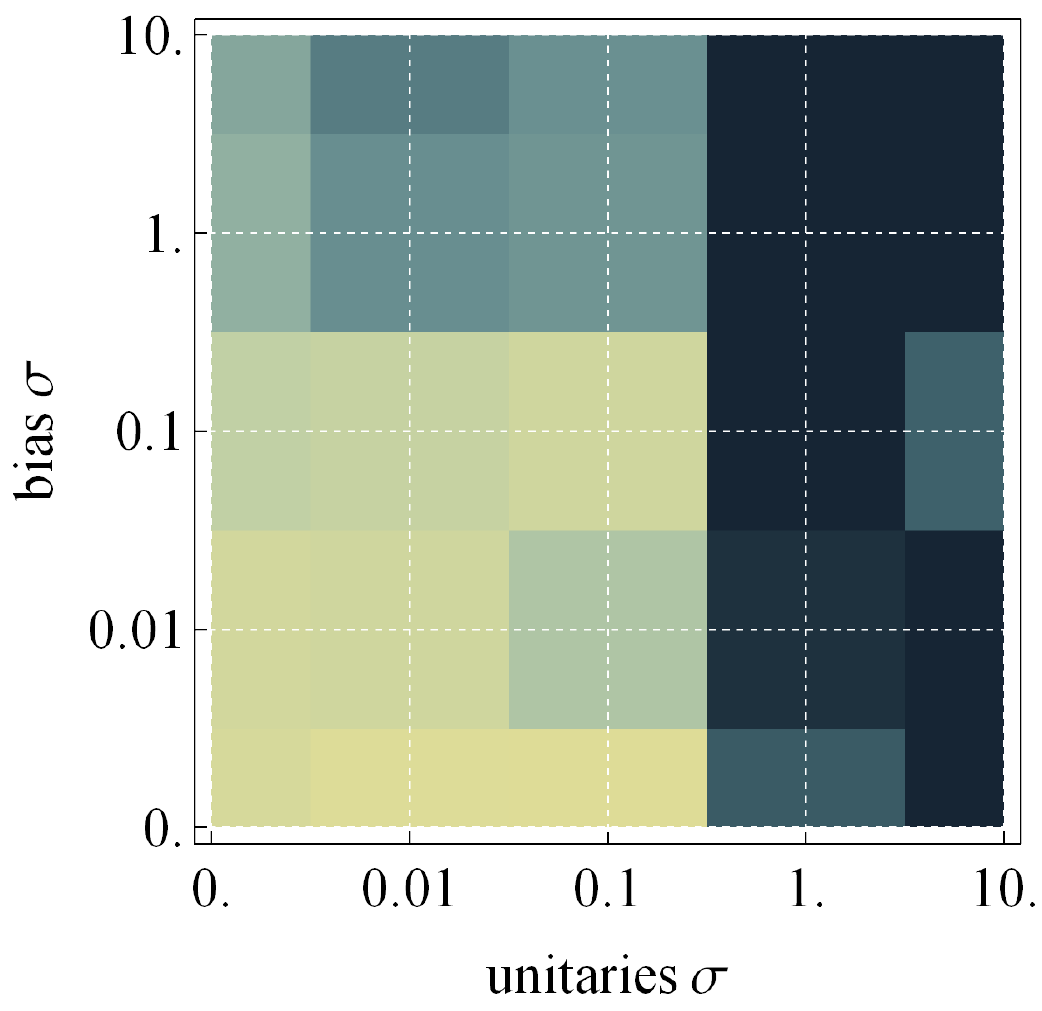}\\
    \tikz{\draw[black,dashed] (0,0) -- (18,0);}\\[5mm]
    \includegraphics[width=.33\textwidth]{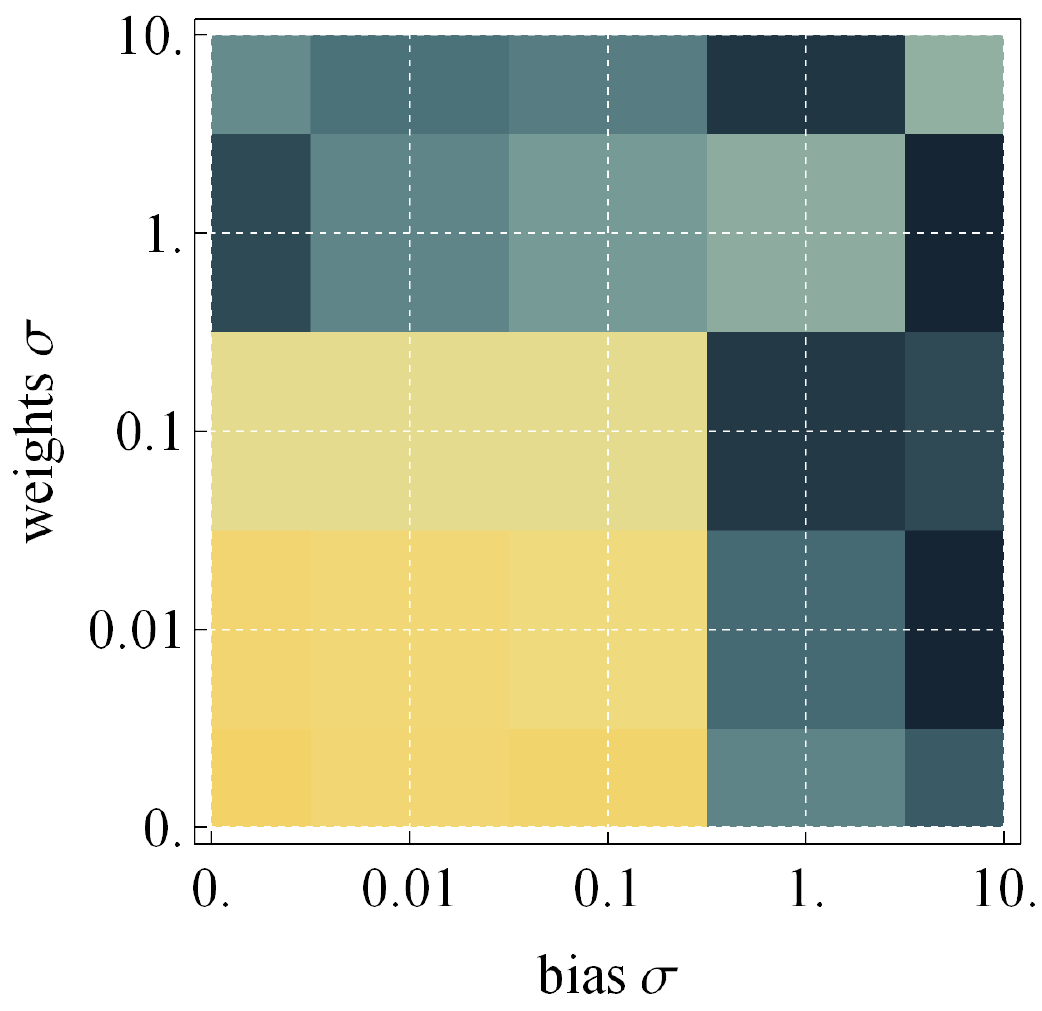}
    \includegraphics[width=.33\textwidth]{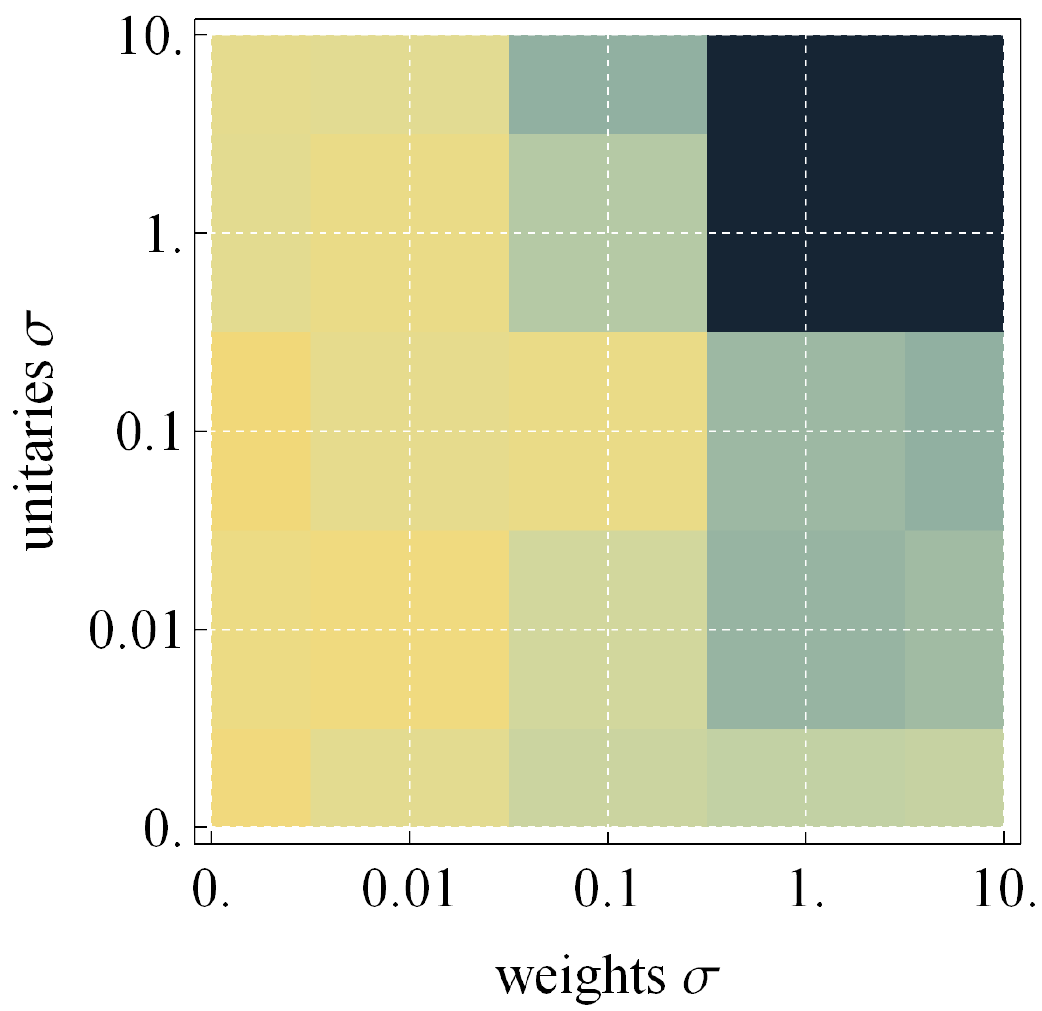}
    \includegraphics[width=.33\textwidth]{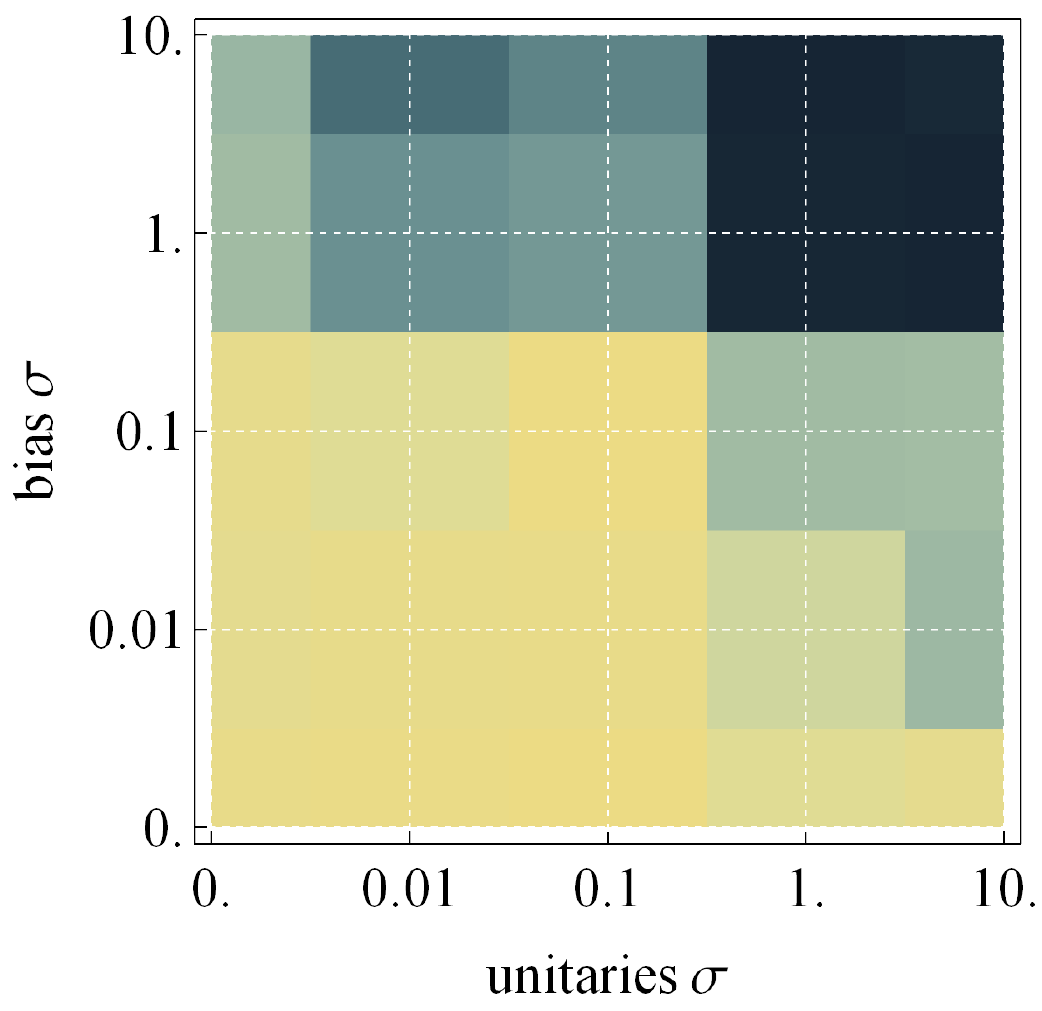}\\
    \raisebox{-2.8mm}{\includegraphics[width=7.5cm]{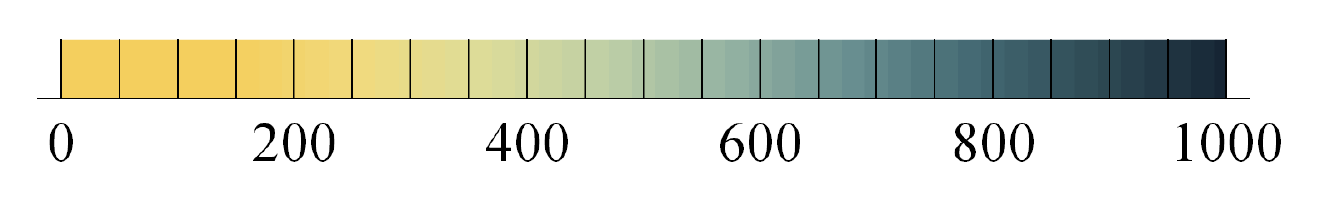}} median number of training steps\\[5mm]
    \end{minipage}
    \caption{Pairwise comparison of median convergence time for hyperparameter initialization of QRNN cell as described in \cref{sec:elman}.
    A cell bias $\mu=\pi/8$ yields the fastest convergence; shown in the third row are the same parameter comparisons as in the second row, but only considering runs with $\mu=\pi/8$.}
    \label{fig:initialization}
\end{figure}

\section{QRNN Postselection Analysis}
A QRNN as described in \cref{sec:model} has two locations where we utilise amplitude amplification with a subsequent measurement to mimic the process of postselection.
This introduces an overhead, since sub-circuits and their inverse operations need to repeated multiple times, depending on the likelihood of the event that we postselect on.
We emphasize that this overhead is only present when running this model on a quantum device; classically, since we have access to the full statevector, we can postselect by multiplying with a projector, and renormalizing the state.
This is also how the quantum neuron is implemented as a pytorch layer.

Amplitude amplification is a generic variant of Grover search, described in detail e.g.\ in \cite{Nielsen2010}.
In brief, a state $\ket{\psi'} = \alpha \ket0 \ket{x} + \sqrt{1-\alpha^2} \ket{1}\ket{y}$ (for normalised $\ket{x},\ket{y}$) has a likelihood $\propto |\alpha|^2$ to be measured in state $\ket0\ket{x}$.
Amplitude amplification allows the state to be manipulated such that this probability can be bumped close to $1$.

More precisely, the variant of amplitude amplification we utilise is called ``fixed-point oblivious amplitude amplification'' \cite{Tacchino2019,Grover2005},
which is suitable for the case where we have a unitary $\op U\ket{\psi} = \ket{\psi'}$ that produces the state (which is where ``oblivious'' comes from); and where we do not know $\alpha$ (which is where ``fixed-point'' comes from).
By repeatedly applying $\op U$ and its inverse $\op U^\dagger$ in a specific fashion, the likelihood of a subsequent measurement to observe outcome $\ket0\ket{x}$ can be amplified to a probability $\ge 1-\epsilon$, with $\BigO(\log\epsilon/|\alpha|)$ many applications of $\op U$.

\subsection{Quantum Neuron}
The first location where amplitude amplification is necessary is in the application of each quantum neuron; the purple meters in \cref{fig:qn-1} in the main text indicate that we would like to measure $\ket0$ on the respective lanes---if a $\ket1$ was measured, a wrong operation results.
As discussed, the authors in \cite{Cao2017} named their quantum neuron with a similar structure a repeat-until-success (RUS) circuit.
Such circuits generally have the feature that the ``recovery'' operation is simple, which essentially means that one can just ``flush and repeat'' the operation until it finally succeeds.
This is true for their first degree quantum neuron, as it is for our higher-degree quantum neuron---but only on the Hilbert space spanned by product states (e.g.\ on inputs like $\ket1\ket0$, but not states like Bell pairs such as $(\ket{00} + \ket{11})/\sqrt 2$).
This means that, without modification, a quantum neuron as proposed in \cite{Cao2017} cannot be lifted to a RUS circuit on the full Hilbert space of input states.
The necessary modification was proposed in \cite{Tacchino2019}: amplify the 0 measurement outcome.
This means that (almost) never encouters the situation where one would have to correct an invalid application of the quantum neuron; as the issue with superposition states only ever occurs when a $1$ is measured, the quantum neuron---when postselected on measuring $0$ every time---works as intended on the full Hilbert space of input states.

Since we can detect failure (i.e.\ measuring 1), we can simply choose the likelihood of measuring 0---i.e.\ $1-\epsilon$---to be such that we do not fail too often; and in case of a failure simply repeat the entire QRNN run. Due to the logarithmic dependence on $\epsilon$ in the time complexity of fixed-point amplitude amplification this is possible with an at most logarithmic overhead in $\epsilon$ and the number of postselections to be done.

But what is the overhead with respect to $\alpha$? I.e.\, when applying a quantum neuron, how many times do we have to apply the neuron and its inverse in order to be able to apply the intended nonlinear transformation in eq.~(2) in the main text?
A loose bound can be readily derived as follows.
The ``good'' overall transformation which we wish to postselect on is a map
\[
\ket0 \longmapsto \cos(\theta)^{2^\mathrm{ord}}\ket0 + \sin(\theta)^{2^\mathrm{ord}}\ket1 \eqqcolon \ket{x}
\quad\text{with}\quad
\left\| \ket{x} \right \|^2 = \cos(\theta)^{2\times 2^\mathrm{ord}} + \sin(\theta)^{2\times 2^\mathrm{ord}}.
\]
As we treat the order $\mathrm{ord}$ of the neuron as a constant (it is a hyperparameter, and it is not beneficial to think about its scaling; choices of $\mathrm{ord} \in \{ 1, 2, 3, 4\}$ seem sensible) is easy to derive $\| \ket{x} \|^2 \ge 1/2^{\mathrm{ord}^2-1}$---which results in an amplitude amplification overhead of about 2, 8, 128, or $32768$ for $\mathrm{ord}=1,2,3,4$, respectively.
For all our experiments we chose $\mathrm{ord}=2$; this choice is based on empirical evidence, and the fact that the activation function for $\mathrm{ord}=2$---shown as the dashed line in Fig.~5 in the main text---features relatively steep slopes around $\theta=\pi/4$ and $3\pi/4$; and a relatively flat plateou around $0$ and $\pi/2$.

\subsection{QRNN Cell Output}
The second point where we amplify is during training.
For each application of the QRNN cell as depicted in \cite{fig:qrnn} in the main text, we write the input bit string onto the in/out lanes with a series of classically-controlled bit flip gates.
After this, a series of stages process the new input together with the hidden cell state.
Each of the gates therein can be \emph{conditioned} on the input, but \emph{do not} change the in/out lane at all (see \cref{fig:qrnn-cell}).
This is crucial: if e.g.\ the input bit string was \texttt{0110}, the overall state of the QRNN after the input has been written is $\ket{0110}\ket{h}$, where $\ket{h}$ represents the hidden state.
The subsequent controlled lanes thus cannot create entanglement between the $\ket{0110}$ state and $\ket{h}$, as $\ket{0110}$ is not in a superposition.
This allows us to reset the in/out lanes with an identical set of bit flips that entered the bit string in first place; resulting in a state $\ket{0000}\ket{h'}$ right at the start of the output stage.
The output neurons can then utilize this clean output state to write an output word, which is measured.

It is this output word that we perform postselection on during training.
For instance, if the character level QRNN is fed an input string (e.g.\ ascii-encoded lower-case English letters) \texttt{fisheries}, then after having fed the network \texttt{fish} the next expected letter is a \texttt{e}.
Yet, at this output stage, all the QRNN does is to present us with a quantum state; measuring the output word results in a distribution over predicted letters, much like in the classical case for RNNs and LSTMs.\footnote{As explained in the main text, depending on whether we run this QRNN on a classical computer or a quantum device, we can either extract these probabilities by calculating the marginal of the statevector---which is done in our pytorch implementation---or by sampling.
The sampling overhead naturally depends on the precision to which one wishes to reproduce the distribution.}
Depending on which outcome is measured, this means a different hidden cell state is retained: if---for our example the state at the end of the output stage in \cref{fig:qrnn-cell}
is
\[
    \ket{\psi} = p_\texttt a \ket{\texttt a}\ket{h_\texttt a} + p_\texttt b \ket{\texttt b}\ket{h_\texttt b} + \ldots + p_\texttt z \ket{\texttt z}\ket{h_\texttt z},
\]
then measuring \texttt{z} collapses the QRNN cell state to $\ket{h_\texttt z}$; measuring \texttt{q} collapses it to $\ket{h_\texttt q}$.

This is a useful feature during inference: if one measures a certain letter, we expect the internal state of the QRNN to reflect this change.\footnote{Note how this change is due to the collapse of an entangled state by simply measuring the output lanes; we never actively modify the cell state.}
This natural source of randomness is e.g.\ the basis for the different handwritten digits produced in \cref{fig:gen-digits}: a measurement of a white pixel determines the likelihood of measuring consecutive white or black pixels further down the line, recreating what is either a \texttt{'0'} or a \texttt{'1'}.

Yet while this collapse of the statevector during inference is a useful feature, during training this results in poor performance, as the output distribution is not predictive enough yet to give any meaningful correlation between measured output and resulting internal state.

To circumvent this, we postselect on the next letter that we expect---e.g.\ in the above example of \texttt{fisheries} we would postselect on finding the letter \texttt{e}.
One point to emphasize here is that it suffices to repeat the \emph{current} QRNN cell unitary (and its inverse) for the amplitude amplification steps; one does not have to iteratively apply the entire QRNN up to that point; the latter would necessarily result in an exponential runtine overhead. This is not the case here.

We chose to analyse the resulting amplitude amplification overhead only empirically, and implemented a monitoring feature into pytorch that allowed us to, at any point in time, track the minimum postselection probability that \emph{would} result in an overhead if running the QRNN on a quantum device.

\begin{figure}[t]
    \includegraphics[width=0.45\textwidth]{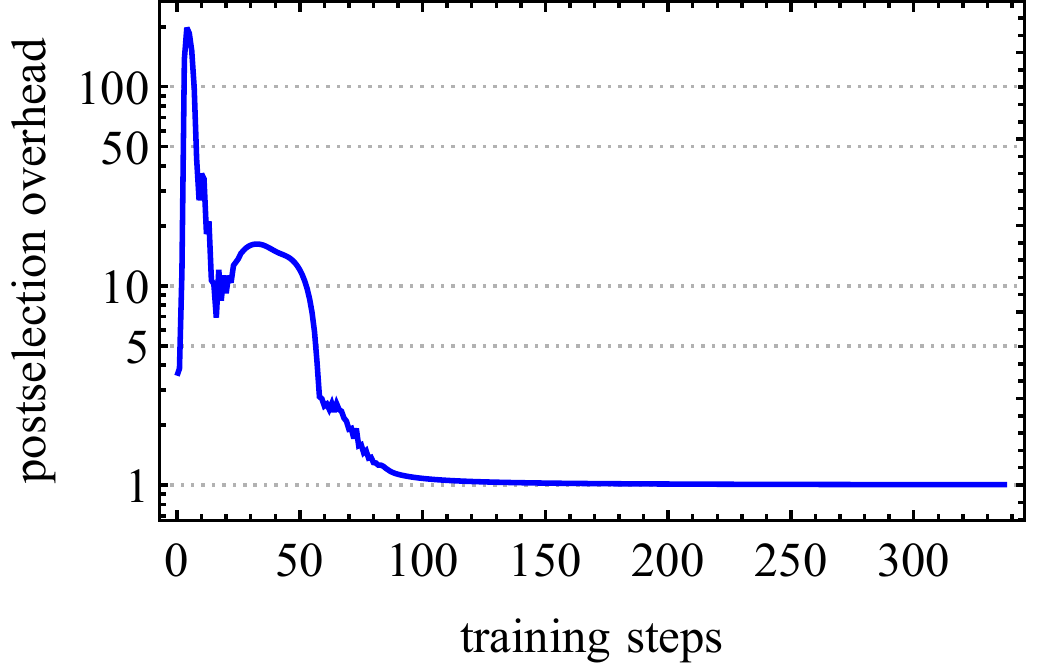}
    \hfill
    \includegraphics[width=0.45\textwidth]{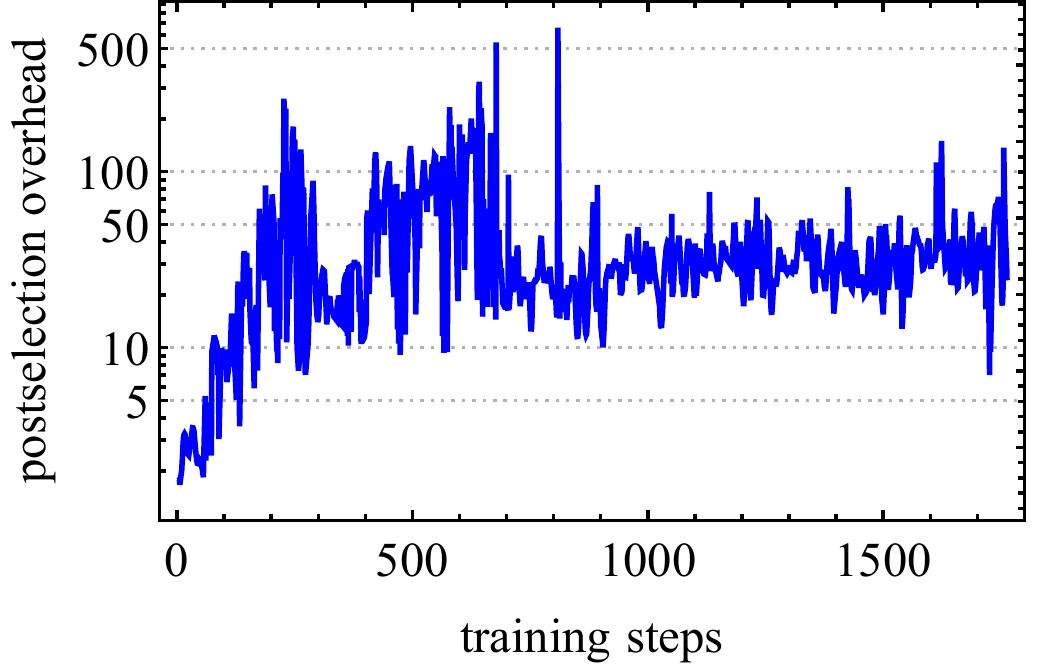}\\
    \includegraphics[width=0.45\textwidth]{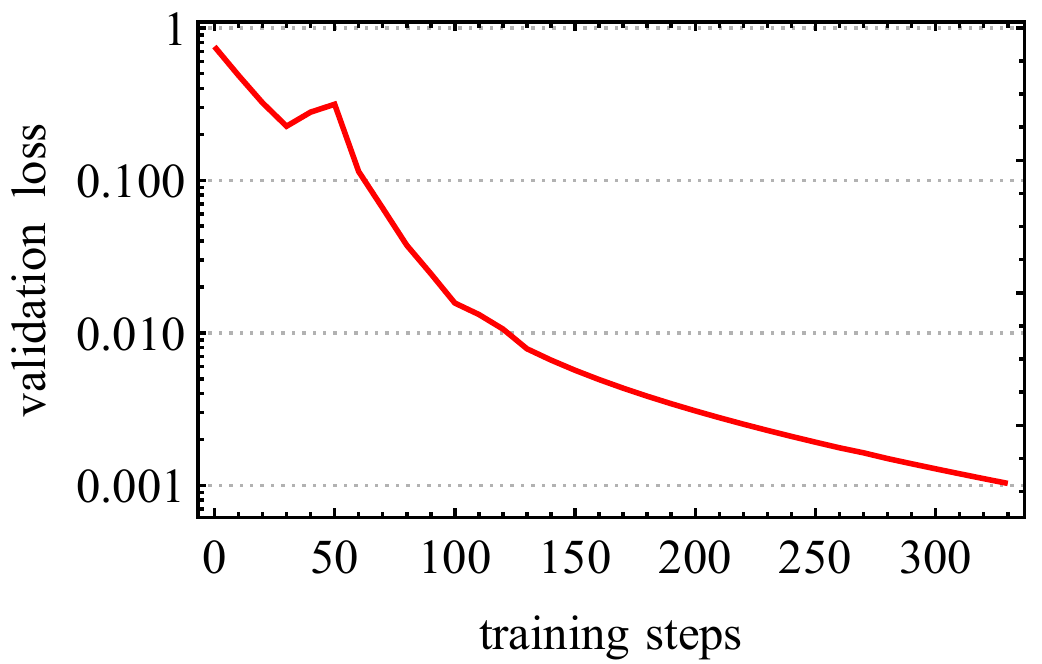}
    \hfill
    \includegraphics[width=0.45\textwidth]{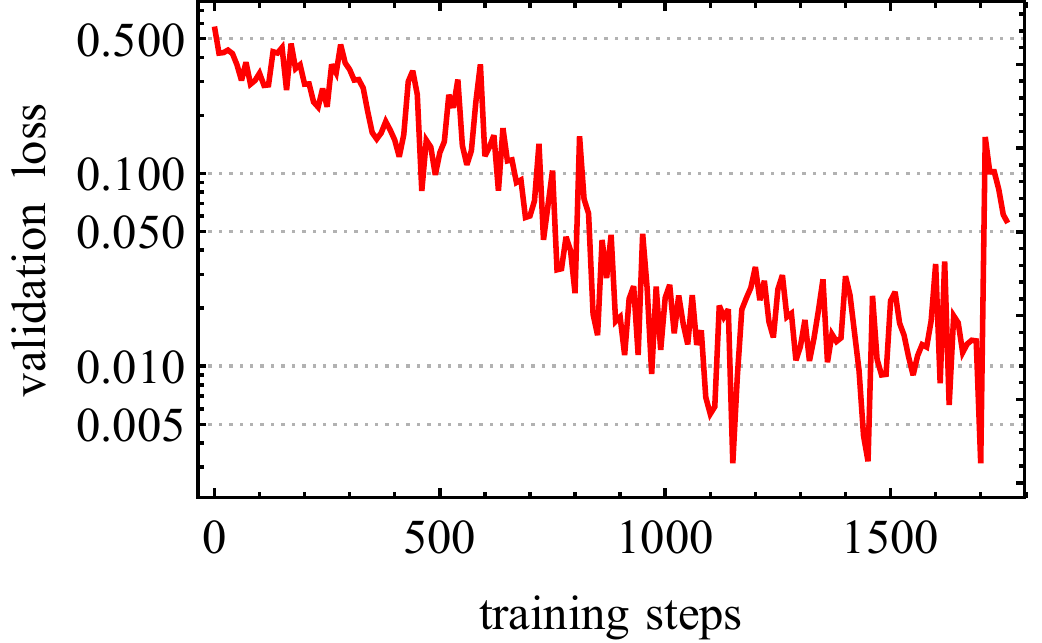}
    \caption{Typical amplitude amplification overhead during training. Left: memorization of simple sequences as described in \cref{sec:memo}; the overhead approaches one as the validation loss converges to zero. Right: pixel-by-pixel MNIST classification from \cref{sec:mnist}; the overhead stabilises around a constant of $\approx 40$ as the validation loss decreases.}
    \label{fig:postselection}
\end{figure}

We found three trends during our experiments.
\begin{enumerate}
    \item The overall postselection overhead was relatively small, but tends to be larger the wider the in/out lanes.
    \item For memorization tasks or learning simple sequences (e.g.\ Elman's XOR test), the overhead started larger, but then converged to one.
    \item For learning more complicated sequences as e.g.\ the pixel-by-pixel MNIST learning task, the postselection probability converged to roughly a constant $>1$.
\end{enumerate}
Examples for the postselection overheads during two representative training tasks are plotted in \cref{fig:postselection}.

\section{QRNN Network Topology}\label{sec:apdx-params}
As explained in Sec.~4.2 in the main text, we used \citeauthor{Elman1990}'s task of learning sequences comprising the three words ``ba'', ``dii'' and ``guuu'' to assess what network topologies work best in this scenario; i.e., we ask the question of how many work stages within the QRNN cell are useful, and what influence the neuron degree\footnote{As a reminder, and as explained in the last section, we set our neurons to have \emph{order} 2. The \emph{degree} of the neuron is the degree of the polynomial of the inputs as shown in eq.~(3) in the main text.}
and workspace size has on the learning speed.

\begin{figure}[t]
    \includegraphics[width=0.33\textwidth]{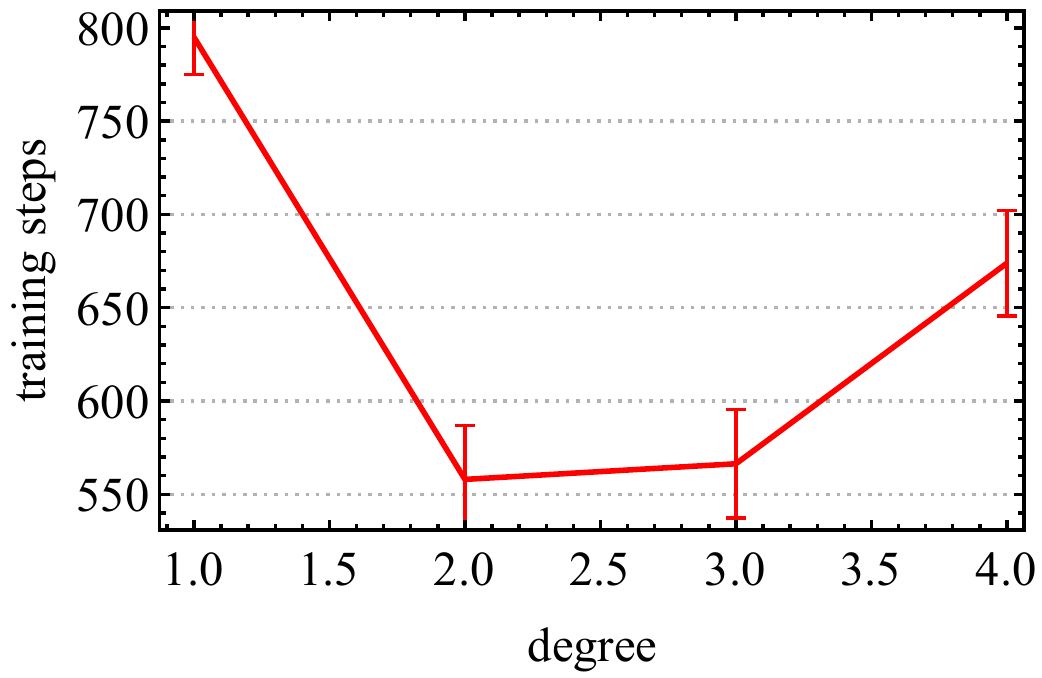}
    \includegraphics[width=0.33\textwidth]{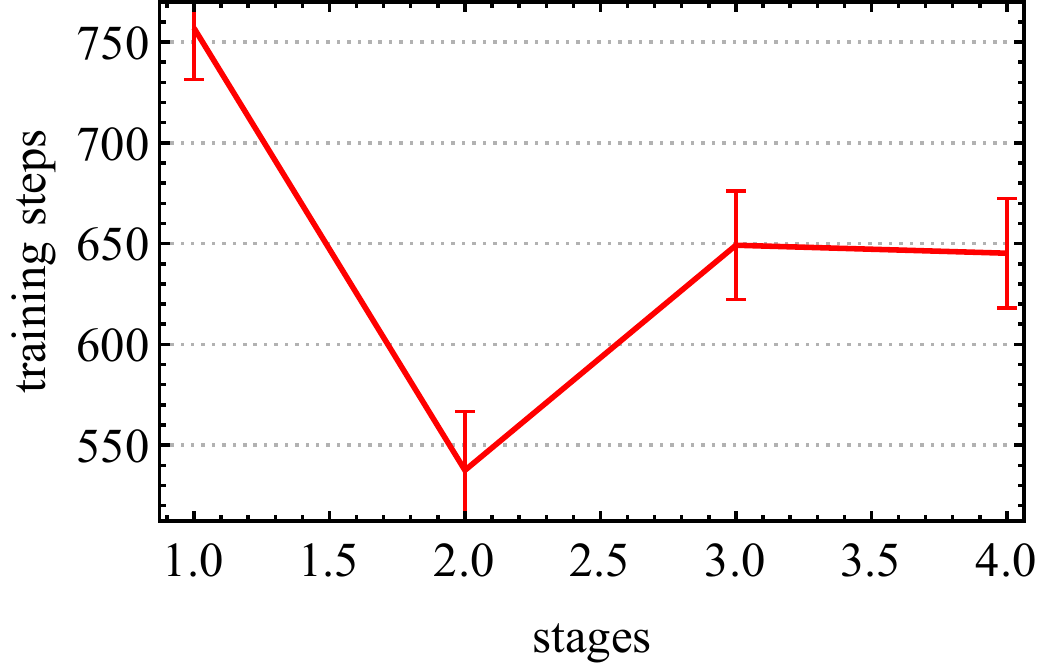}
    \includegraphics[width=0.33\textwidth]{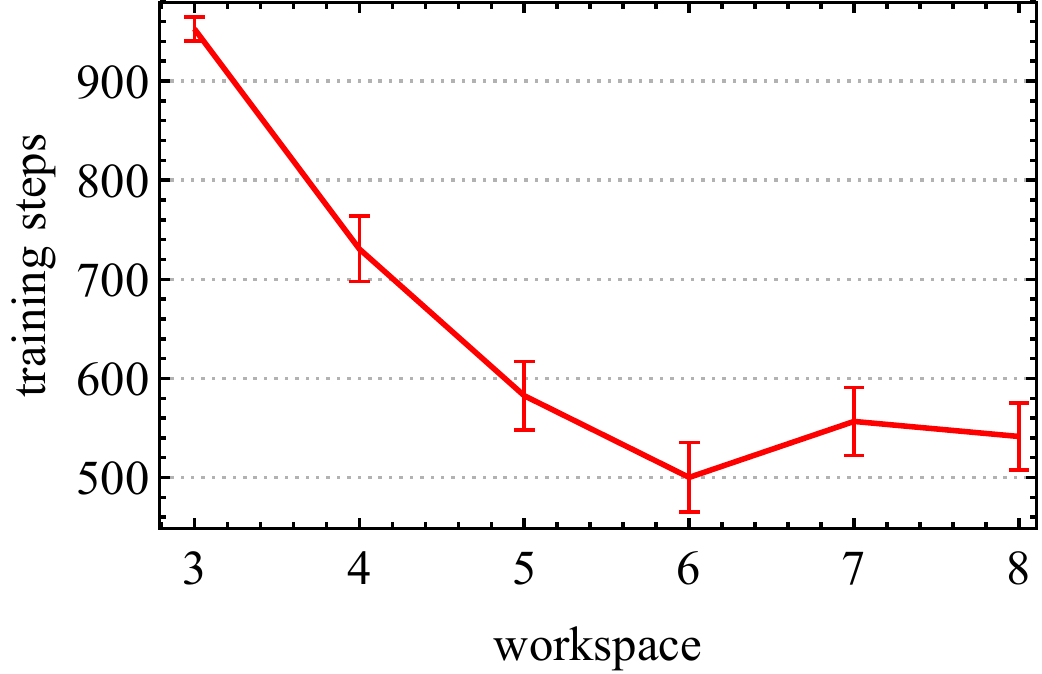}
    \caption{Average number of training steps for sentence learning task described in \cref{sec:memo}, for various combinations of neuron degree, neuron stages, and workspace.
    For this task, we found that a combination of degree 2, workspace 6, and 2 stages performed best.}
    \label{fig:topology}
\end{figure}

Our findings are summarised in \cref{fig:topology}.
The input for this task has a width of three bits (which suffices for the six different letters used), so the useful degree in the input stage in the QRNN cell is upper-bound by three.
A higher degree becomes useful only if the workspace size is increased accordingly.

Despite this, we found that a degree of two is already optimal; a degree of three is not better, and a degree of four has a longer expected convergence time again.
This is likely due to the larger number of parameters necessary for higher-degree neurons, as explained in Fig.~2 in the main text.

A similar picture can be seen when looking at the number of work stages in the QRNN cell: a single stage takes considerably longer than two stages; for more stages, the learning time increases again.

In contrast to this, it appears that the more workspace we have present the better; yet even here there appears to be a plateau when going to $\ge 6$ qubits.
This is likely due to the simplicity of the learning task.
For instance, as listed in Tab.~1 in the main text, a workspace of six was enough to classify MNIST when using data augmentation (which, with an input width of 2 bits, and an order 2 neuron requires two ancillas, resulting in 10 qubits overall).
On the other hand, the pixel-by-pixel task required a higher information capacity; we found that a workspace of eight performed better in this setting.

So in general, and as in the case of classical neural networks, there must be a tradeoff between the number of parameters and the expected learning time.
Too few parameters and the model does not converge.
Too many parameters become costly, and potentially start to overfit the dataset.
While the QRNN workspace size has a direct analogy to layer width in classical RNNs and other network architectures, and stages with the depth of the network, the quantum neuron's degree finds no good analogy in common neural network architectures.

\end{document}